\documentclass{article}

\PassOptionsToPackage{numbers, compress}{natbib}

\usepackage[final]{neurips_2025}

\usepackage[utf8]{inputenc} 
\usepackage[T1]{fontenc}    
\usepackage{hyperref}       
\usepackage{url}            
\usepackage{booktabs}       
\usepackage{amsfonts}       
\usepackage{nicefrac}       
\usepackage{microtype}      
\usepackage{xcolor}         

\usepackage[ruled, vlined, linesnumbered]{algorithm2e}
\usepackage{amsmath,amsthm,amssymb}
\usepackage{bbm}
\usepackage{bm}
\usepackage[capitalize]{cleveref}
\usepackage{enumitem}
\usepackage{graphicx}
\usepackage{lipsum}  
\usepackage{multirow}
\usepackage{pgfplots}
\usepackage{placeins}
\usepackage{subcaption}
\usepackage{soul}
\usepackage{tabularx}
\usepackage{wrapfig}
\usepackage{xspace}

\usepackage{listings}
\usepgfplotslibrary{groupplots,dateplot}
\usetikzlibrary{patterns,shapes.arrows}
\pgfplotsset{compat=newest}

\usepackage{tcolorbox}
\newtcolorbox{prompt}[1]{colback=gray!20,colframe=gray!50!black,fonttitle=\bfseries,title=#1}

\newcommand{\hrpo}{{HRPO}\xspace}
\newcommand{\hrpofull}{{hybrid reasoning policy optimization}\xspace}
\newcommand{\HrpoFull}{{Hybrid Reasoning Policy Optimization}\xspace}

\title{Hybrid Latent Reasoning via Reinforcement Learning}

\author{%
  Zhenrui Yue$^1$, Bowen Jin$^1$, Huimin Zeng$^1$, Honglei Zhuang$^2$, Zhen Qin$^2$, Jinsung Yoon$^2$, \\
  \textbf{Lanyu Shang$^3$, Jiawei Han$^1$, Dong Wang$^1$} \\
  $^1$University of Illinois Urbana-Champaign, $^2$Google, $^3$LMU \\
  \texttt{\{zhenrui3,bowenj4,huiminz3,lshang3,hanj,dwang24\}@illinois.edu,} \\
  \texttt{\{hlz,zhenqin,jinsungyoon\}@google.com, lanyu.shang@lmu.edu} \\
}

\begin{document}

\maketitle

\begin{abstract}
Recent advances in large language models (LLMs) have introduced latent reasoning as a promising alternative to autoregressive reasoning. By performing internal computation with hidden states from previous steps, latent reasoning benefit from more informative features rather than sampling a discrete chain-of-thought (CoT) path. Yet latent reasoning approaches are often incompatible with LLMs, as their continuous paradigm conflicts with the discrete nature of autoregressive generation. Moreover, these methods rely on CoT traces for training and thus fail to exploit the inherent reasoning patterns of LLMs. In this work, we explore latent reasoning by leveraging the intrinsic capabilities of LLMs via reinforcement learning (RL). To this end, we introduce \hrpofull (\hrpo), an RL-based hybrid latent reasoning approach that (1)~integrates prior hidden states into sampled tokens with a learnable gating mechanism, and (2)~initializes training with predominantly token embeddings while progressively incorporating more hidden features. This design maintains LLMs' generative capabilities and incentivizes hybrid reasoning using both discrete and continuous representations. In addition, the hybrid \hrpo introduces stochasticity into latent reasoning via token sampling, thereby enabling RL-based optimization without requiring CoT trajectories. Extensive evaluations across diverse benchmarks show that \hrpo outperforms prior methods in both knowledge- and reasoning-intensive tasks. Furthermore, \hrpo-trained LLMs remain interpretable and exhibit intriguing behaviors like cross-lingual patterns and shorter completion lengths, highlighting the potential of our RL-based approach and offer insights for future work in latent reasoning.
\end{abstract}

\section{Introduction}

\begin{figure}[t]
    \centering
    \includegraphics[trim=0.3cm 2.8cm 0.6cm 3.2cm, clip, width=\textwidth]{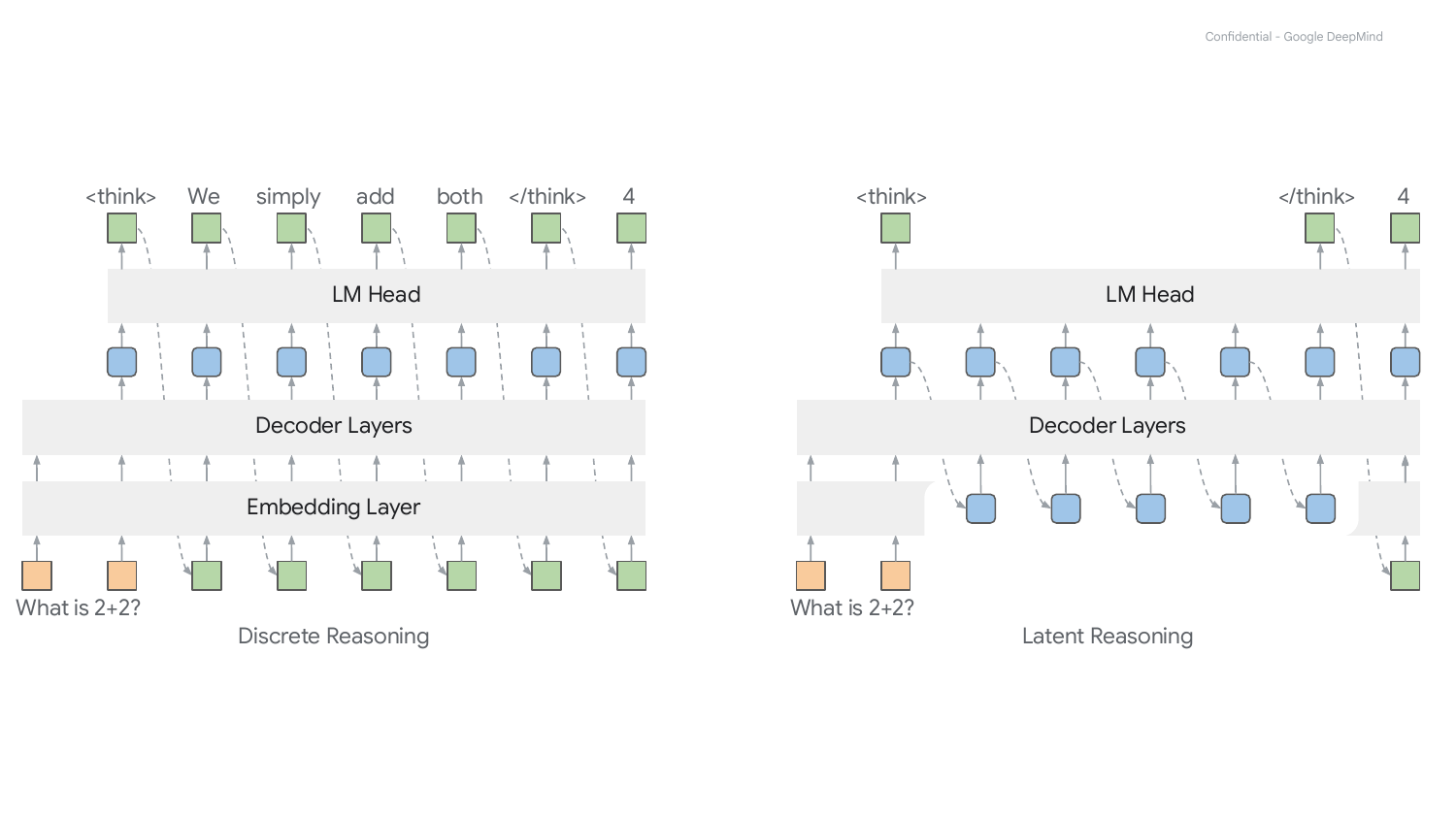}
    \caption{Comparison between discrete reasoning (left) and latent reasoning (right). Unlike the autoregressive sampling process in discrete reasoning, latent reasoning incorporates hidden representations from previous steps to enhance reasoning performance (between \texttt{<think>} and \texttt{</think>}).}
    \label{fig:intro}
\end{figure}

Latent reasoning has emerged as a compelling alternative to traditional autoregressive reasoning methods in large language models (LLMs)~\cite{geiping2025scaling, shen2025codi, tack2025llm}. In contrast to the conventional chain-of-thought (CoT)~\cite{wei2022chain, jaech2024openai, guo2025deepseek}, which relies on the discrete decoding and sampling process, latent reasoning enables LLMs to reason internally with continuous hidden representations from the previous steps. For instance, Coconut~\cite{hao2024training} achieves latent reasoning by utilizing the model's last hidden state as `continuous thought', feeding it back as input embeddings to the next reasoning step, thereby matching the performance of CoT on reasoning-intensive tasks. To show the difference between the autoregressive generation and latent reasoning, we compare both approaches in \Cref{fig:intro}.

Nevertheless, existing methods in latent reasoning utilize extensive CoT traces for training. That is, CoT trajectories are required to learn informative latent representations. An example is CODI~\cite{shen2025codi}, which adopts self-distillation to train on discrete CoT tokens and transfers learnt features into continuous thoughts. Although recurrent latent reasoning removes the need for CoT data, it relies on training a multi-block LLM from scratch to reason internally~\cite{geiping2025scaling}. Moreover, these methods employ tailored training paradigms for latent representation learning, incurring high training costs and overlooking the inherent reasoning capabilities of LLMs~\cite{hao2024training, geiping2025scaling, shen2025efficient}. For example, Coconut~\cite{hao2024training} requires multi-stage training on CoT steps, which not only increases training compute but also delays the model's acquisition of complete reasoning chains~\cite{shen2025codi}. Furthermore, we find that latent reasoning is often incompatible with LLMs due to the discrepancy between output hidden states and input embeddings (as we show \Cref{sec:analysis}). That is, feeding hidden states into the next decoding step degrades generation quality (e.g., repetition, incoherence), causing difficulties in adapting LLMs for latent reasoning. Therefore, an ideal latent reasoning method should capitalize on pretrained LLMs’ generalizability by seamlessly integrating continuous representations, preserving LLMs' interpretability while mitigating CoT‐dependent extensive training for broader applicability.

To this end, we introduce \hrpofull (\hrpo), a novel hybrid latent reasoning optimization framework based on reinforcement learning (RL). \hrpo unifies policy learning with latent reasoning, thereby utilizing the LLMs' intrinsic reasoning patterns without relying on CoT trajectories. To preserve the generative capabilities while encouraging the model to reason in the continuous space, \hrpo introduces a gating mechanism to gradually incorporate hidden state representations from previous steps into sampled token embeddings. The gating mechanism is initially configured in a way that the inputs come predominantly from the sampled tokens. As training progresses, the gate learns to incorporate richer, more informative features from previous hidden states for improved internal reasoning. Since the sampling operation introduces stochasticity, \hrpo rollouts can be performed like standard RL methods, with hybrid outputs (tokens and latent representations) stored in the rollout buffer for policy updates. For optimization, \hrpo leverages a simple outcome-based reward and employs the hybrid rollout buffer to calculate log probabilities, enabling policy gradient updates that adaptively integrate both token-level and latent representations. By bridging discrete and continuous reasoning, \hrpo provides a scalable and training-efficient solution that unlocks latent reasoning in existing LLMs. As a result, \hrpo enhances the adaptability of latent reasoning and leads to superior performance on both knowledge- and reasoning-intensive tasks. We highlight our contributions in the following\footnote{Our implementation is available at https://github.com/Yueeeeeeee/\hrpo.}: 
\begin{itemize}
    \item We introduce \hrpo, the first reinforcement learning-based approach for hybrid reasoning, empowering LLMs to autonomously develop latent reasoning capabilities.
    \item We design a gating mechanism to preserve LLMs' generative abilities, which starts by prioritizing sampled token embeddings and, through RL-driven updates, progressively incorporates the continuous representations.
    \item By leveraging the LLMs' inherent reasoning patterns through \hrpo, we mitigate the need for chain-of-thought annotations and expensive multi-stage training, offering an efficient and scalable alternative to existing latent reasoning methods.
    \item To show the efficacy of the proposed hybrid latent reasoning, we evaluate on multiple knowledge and reasoning benchmarks and show that it outperforms existing models and latent reasoning baselines, demonstrating consistent performance gains across diverse scenarios. In addition, we provide insights into RL-based training of latent reasoning models and present intriguing reasoning patterns emerging from \hrpo.
\end{itemize}
\section{Related Work}

\subsection{Latent Reasoning}
Early research in latent reasoning focuses on analyzing the latent space computation within transformer models~\cite{biran2024hopping, yang2024large}. For example, \citet{biran2024hopping} study multi-hop reasoning and show that `back-patch' features from later layers can improve performance on challenging queries. Alternatively, latent representations can be used to construct informative features as in-context demonstrations to enhance few-shot performance at test-time~\cite{xu2024lars, zhuang2024vector}. In particular, \citet{xu2024lars} exploit latent skills to select in-context examples for reasoning-intensive tasks. Different from this line of work, hidden reasoning is also proposed to improve generative capabilities by incorporating latent variables into language modeling~\cite{geiping2025scaling, kong2025scalable}. For instance, \citet{geiping2025scaling} propose a depth-recurrence language model that injects latent variables and iteratively processes them to derive the final states used for decoding. Similarly, special tokens (e.g. \texttt{<pause>}) are inserted to allocate extra test-time compute for internal reasoning, leading to improvements across diverse scenarios~\cite{goyal2023think, pfau2024let}. \citet{pfau2024let} argue that filler tokens act as intermediate reasoning steps in multi-token computations, yielding measurable performance gains on parallelizable problems. Furthermore, implicit reasoning methods transform explicit, token-level reasoning trajectories into internal reasoning to enhance efficiency or accuracy~\cite{deng2023implicit, deng2024explicit}. For instance, CODI~\cite{shen2025codi} employs a self-distillation to framework to align explicit and implicit reasoning tokens for improved performance. Concurrent to our work, hidden reasoning approaches~\cite{hao2024training, shen2025efficient, su2025token} leverage previous output hidden states as next input embeddings, enabling compact yet informative internal reasoning. Nonetheless, the majority of existing methods require processed traces and extensive training. In contrast, we focus on hybrid latent reasoning through reinforcement learning to exploit the inherent generation capabilities of LLMs.

\subsection{Reinforcement Learning}
Reinforcement learning (RL) is a paradigm where an agent interacts with an environment, receives feedback, and learns to make decisions that maximize cumulative rewards over time~\cite{sutton1998reinforcement}. Recently, RL has been introduced to improve language models by learning from implicit human feedback (RLHF)~\cite{ouyang2022training}. Such fine-tuning typically employs policy gradient algorithms and their variants like REINFORCE~\cite{sutton1999policy}. To reduce variance, actor-critic methods like A2C~\cite{mnih2016asynchronous} are proposed to compute a learnt baseline and leverage advantage estimates for better training dynamics. Similarly, proximal policy optimization (PPO)~\cite{schulman2017proximal} introduces a clipped surrogate objective to bound policy updates, thereby achieving training stability and robustness to hyperparameter choices. Parallel to these approaches, direct preference optimization (DPO)~\cite{rafailov2023direct} is introduced to directly optimize language models using pairwise human preference comparisons. DPO's simpler variant such as SimPO~\cite{meng2024simpo} further mitigates the need of reference models. Despite DPO's efficiency, online RL methods remain preferred for their consistent superior performance~\cite{xu2024dpo}. Recently, reinforce leave-one-out (RLOO)~\cite{ahmadian2024back} proposes REINFORCE-style RL that generates multiple responses and utilizes the mean reward of the other responses as a baseline. Similarly, group relative policy optimization (GRPO)~\cite{shao2024deepseekmath} and REINFORCE++~\cite{hu2025reinforce++} compute baselines from group-level or batch-level reward scores across candidate completions, and thus reduce memory overhead while maintaining accuracy and stability for complex tasks. In this work, we design a novel online RL–driven approach to incentivize hybrid latent reasoning by progressively incorporating hidden states into LLM inputs, thereby providing richer representations for improved reasoning performance.
\section{Methodology}
\label{sec:method}

\begin{figure}[t]
    \centering
    \includegraphics[trim=0 2.4cm 0.1cm 2.4cm, clip, width=\textwidth]{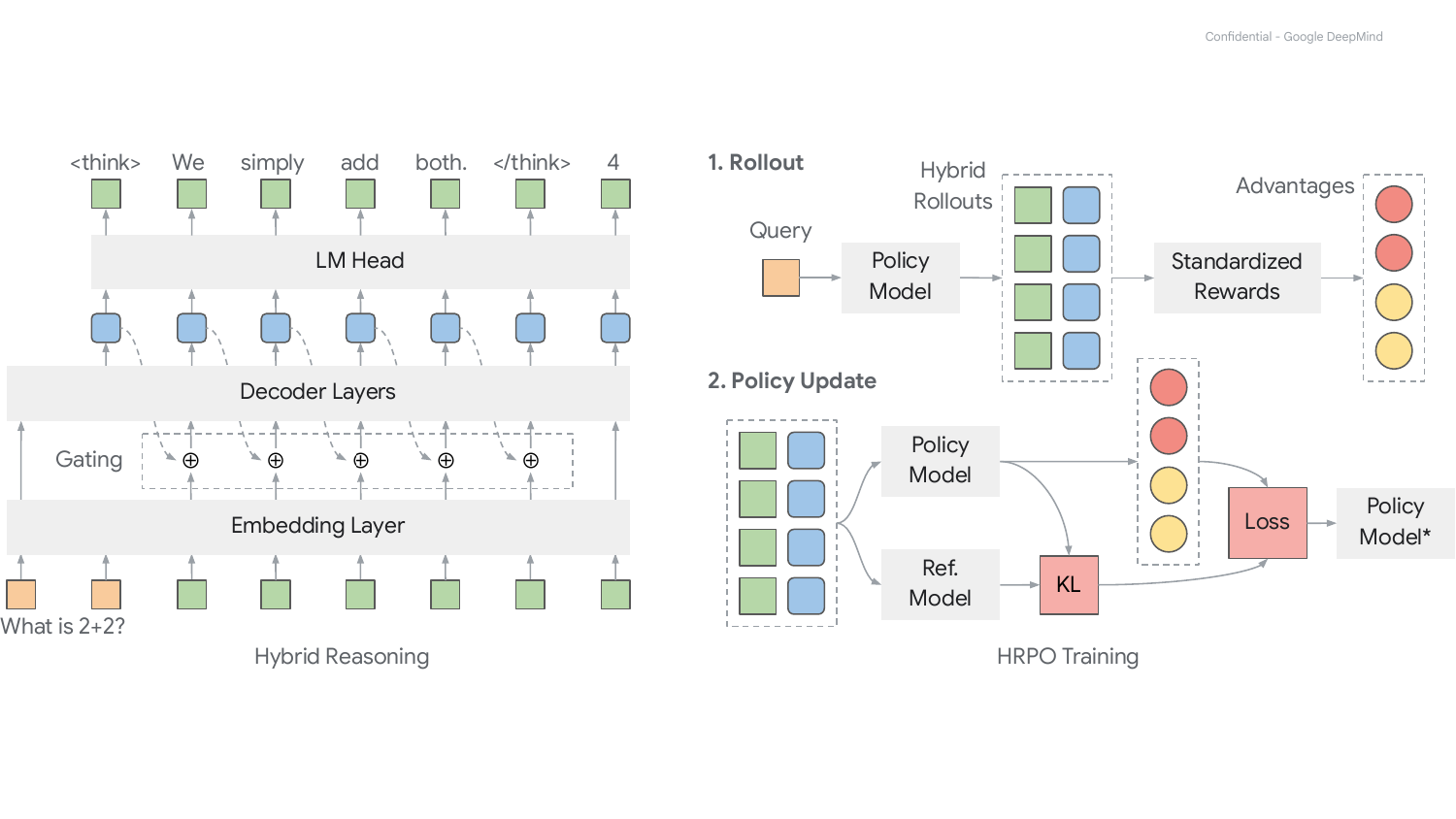}
    \caption{Hybrid reasoning with gating (left) and \hrpofull (right). During rollouts, the reasoning trajectory is generated hybridly with both discrete tokens and latent features, and for policy update, we compute the \hrpo loss using the hybrid rollout buffer to update the model.}
    \label{fig:method}
\end{figure}

\subsection{Hybrid Reasoning with Gating}
We first describe our notation and settings for hybrid latent reasoning. For input query $x = [x_1, x_2, \ldots, x_t]$ and its corresponding token embeddings $E = [e_1, e_2, \ldots, e_t]$, we describe the raw hidden states from the LLM output at step $t$ with $\hat{h}_t$, namely:
\begin{equation}
    \hat{H} = [\hat{h}_1,\hat{h}_2,\ldots,\hat{h}_t] = \mathtt{Transformer}(E),
\end{equation}
in which \texttt{Transformer} denotes the transformer model (i.e., decoder layers), $\hat{H}$ represents the final‐layer hidden states produced by the \texttt{Transformer}. With the LM head (\texttt{Head}), the next output token $\hat{x}_{t+1}$ can be sampled from the output distribution over the vocabulary via:
\begin{equation}
    \hat{x}_{t+1} \sim \texttt{softmax}(\mathtt{Head}(\hat{h}_t)).
\end{equation}
However, hidden states often lie outside the model's token embedding manifold, which degrades generation quality when fed directly. To avoid this, we project $\hat{h}_t$ back into the embedding space to ensure the inputs conform to the model's learned distribution. Specifically, we use the output probabilities $p_{t+1}$ to compute a weighted interpolation over the vocabulary:
\begin{equation}
    h_{t+1} = W_{e}^T \frac{p_{t+1}}{\|p_{t+1}\|}, \quad \mathrm{with} \quad p_{t+1} = \texttt{softmax}(\frac{\texttt{Head}(\hat{h}_t)}{\tau}),
    \label{eq:hidden_states}
\end{equation}
in which $\tau$ is the temperature and $W_{e}$ denotes the embedding matrix of the LLM. In other words, we compute the next input embedding as a weighted sum of all token embeddings, with weights given by $p_{t+1}$. In addition, $p_{t+1}$ is normalized to preserve the scale and variance of the output vector. This sampling-free mapping ensures differentiability and aligns the projected embedding with the model's native input space, thus leading to improved training dynamics (see \Cref{sec:analysis}).

While interpolated embeddings preserve semantic continuity, directly feeding $h_{t+1}$ as the next token input removes stochasticity and injects noise from irrelevant tokens, causing degraded generation within RL rollouts. As such, we design a hybrid approach for latent reasoning by gradually imposing hidden state representations into the sampled token embeddings with a gating mechanism. Drawing on gated recurrence models~\cite{de2024griffin, orvieto2023resurrecting}, we formulate the gating mechanism as:
\begin{equation}
\begin{aligned}
    r_t & = \sigma (W_a \hat{e}_{t+1} + b_a), \\
    i_t & = \sigma (W_x \hat{e}_{t+1} + b_x), \\
    a_t &= \texttt{exp}(-c \cdot \texttt{softplus}(\Lambda) \odot r_t), \\
    e_{t+1} & =
    \left\{ 
        \begin{array}{lc}
            a_t \odot \hat{e}_{t+1} + \sqrt{1-a_t^2} \odot (i_t \odot h_{t+1}) & t \in \texttt{think}, \\
            \hat{e}_{t+1} & t \not\in \texttt{think}, \\
        \end{array}
    \right.
    \label{eq:thinking_residual}
\end{aligned}
\end{equation}
$e_{t+1}$ is the resulting hybrid input for the next step, $\hat{e}_{t+1}$ denotes the embedding of the sampled discrete token $\hat{x}_{t+1}$, whereas $h_{t+1}$ is the projected hidden states as in \Cref{eq:hidden_states}. The gates $r_t$ and $i_t$ leverages sigmoid function $\sigma$ to control the blending, $a_t$ scales $\hat{e}_{t+1}$, $c$ is a fixed scaling constant, and $\Lambda$ is a learnable vector. Note that hybrid reasoning only applies during the reasoning phase (i.e., $t \in \texttt{think}$), while the final answer is still generated via standard autoregressive decoding, as we show in \Cref{fig:method} (left). By initializing $a_t \rightarrow 1$ (see \Cref{sec:implementation}), the inputs first draw predominantly from the sampled token embeddings, thereby effectively preserving the LLM's generative capabilities. As the training progresses, the value range of $a_t$ converges to an optimum range and thus incorporates informative features from both hidden representations and sampled tokens. 

Overall, our hybrid reasoning approach projects hidden states into the embedding space via weighted interpolation. Moreover, the sampling steps preserve stochasticity for effective reinforcement learning. We employ a plug-and-play gating mechanism that initially prioritizes sampled token embeddings while gradually integrating latent signals, providing richer inputs for subsequent reasoning.

\subsection{\HrpoFull (\hrpo)}
Rather than relying on strong supervision, we optimize the policy model via hybrid rollouts using reinforcement learning (RL), fully harnessing LLMs' native reasoning capabilities. Inspired by recent RL advances such as group relative policy optimization (GRPO)~\cite{shao2024deepseekmath}, we introduce \hrpofull (\hrpo), an efficient RL-driven framework that enable LLMs to fuse discrete tokens with continuous representations for hybrid reasoning.

As illustrated in \Cref{fig:method} (right), the proposed \hrpo optimizes the policy (parameterized by $\theta$) to maximize the expected reward for input $x$ drawn from dataset $\mathcal{D}$ and the sampled hybrid outputs $y$ (discrete tokens) and $H$ (hidden representations):
\begin{equation}
   \max_{\theta} \mathbb{E}_{(x, y) \sim \mathcal{D}, (\hat{y}, H) \sim \pi_{\theta}(\cdot|x)} [r(a, y)],
\end{equation}
where $r$ is a simple outcome-based reward function and $a$ denotes the ground truth answer (i.e., it outputs 1 for correct prediction in $y$ and 0 otherwise). The rewards are computed solely on the discrete tokens within the answer span. To obtain an unbiased, low-variance advantage for hybrid latent reasoning, we generate $g$ hybrid rollouts per input query and compute the advantages by standardizing the rewards within the group (i.e., for the $i$-th response, the advantage is calculated by $\hat{A}_i = \frac{r_i - \texttt{mean}([r_1, r_2, \ldots, r_g])}{\texttt{std}([r_1, r_2, \ldots, r_g])}$). Consequently, the policy gradients can be estimated with:
\begin{equation}
\begin{aligned}
\nabla_{\theta} \mathcal{J}_{\mathrm{\hrpo}}(\theta) &= \mathbb{E}_{x \sim \mathcal{D}, \{ (y_i, H_i) \}_{i=1}^g \sim \pi_{\theta}(\cdot|x)} \\
& \left[ \frac{1}{g} \sum_{i=1}^{g} \frac{1}{|y_i|} \sum_{t=1}^{|y_i|} \nabla_{\theta} \log \pi_{\theta}(y_{i, t} | x, y_{i,<t}, H_{i,<t}) \hat{A}_{i, t} \right] -\beta \nabla_{\theta} \mathbb{D}_{K L}[\pi_\theta \| \pi_{\mathrm{ref}}],
\label{eq:hrpo}
\end{aligned}
\end{equation}
where $\pi_{\mathrm{ref}}$ denotes the reference model and KL-divergence acts as a regularizer, controlled by hyperparameter $\beta$. This objective follows a simple REINFORCE‐style formulation, fusing discrete token inputs with continuous hidden representations across the reasoning span via the introduced gating mechanism. The hybrid trajectories that yield higher returns are assigned larger advantage estimates, encouraging policy updates to increase the log probabilities of their subsequent reasoning tokens. For the KL divergence term, we compute log probabilities using solely token IDs for $\pi_{\mathrm{ref}}$, as we find it more effective in preserving training stability. Different from PPO~/~GRPO objectives, we omit the likelihood ratio and directly use raw log probabilities in \Cref{eq:hrpo} because ratio clipping is rarely encountered under our conservative learning schedule. Furthermore, since the hidden representations are directly tied to the parameters $\theta$, each trajectory should only be used for a single gradient update; attempting to reuse it—even with importance sampling—violates the on-policy constraints. As such, our \hrpo implementation remains lightweight, strictly on-policy and could be seamlessly combined with further RL optimizations.

In summary, the proposed \hrpo framework unifies hybrid latent reasoning under a simple RL objective that fully leverages LLMs' intrinsic reasoning capabilities. During rollouts, the decoding process progressively fuses discrete and continuous representations through a learnable gate, preserving coherence while exploiting hidden states. For policy updates, \hrpo derives advantages directly from outcome rewards and performs policy gradient steps with KL regularization. As a result, \hrpo incentivizes LLMs to dynamically integrate sampled tokens with latent representations, delivering stable and efficient on-policy hybrid reasoning training without a separate value function.
\section{Experiments}
\label{sec:exp}

We evaluate \hrpo on both knowledge- and reasoning-intensive tasks: (1)~open-domain \& multi-hop knowledge-intensive question answering (Knowledge); and (2)~science, technology, engineering or mathematics (STEM) benchmarks. The experimental results are reported as follows.

\begin{table}[t]
\centering
\caption{Evaluation performance of various larger LLMs and trained models on open-domain and multi-hop QA benchmarks. The table reports exact match scores based on top-$3$ retrieved documents on five datasets: NQ, TriviaQA, HotpotQA, 2WikiMQA and Bamboogle. The upper block reports results for several RAG baselines using the larger Qwen 2.5 7B LLM, while the lower two blocks evaluate smaller Qwen models (1.5B and 3B) trained with different strategies.}
\begin{tabular}{@{}lcccccc@{}}
\toprule
             & NQ             & TriviaQA       & HotpotQA       & 2WikiMQA       & Bamboogle      & Average        \\ \midrule
\multicolumn{7}{c}{Qwen2.5-7B-Instruct}                                                                            \\ \midrule
QA           & 0.134          & 0.408          & 0.183          & \textbf{0.250} & 0.120          & 0.219          \\
CoT          & 0.048          & 0.185          & 0.092          & 0.111          & 0.232          & 0.134          \\
IRCoT        & 0.224          & 0.478          & 0.133          & 0.149          & 0.224          & 0.242          \\
Search-o1    & 0.151          & 0.443          & 0.187          & 0.176          & \textbf{0.296} & 0.251          \\
RAG          & \textbf{0.349} & \textbf{0.585} & \textbf{0.299} & 0.235          & 0.208          & \textbf{0.335} \\ \midrule
\multicolumn{7}{c}{Qwen2.5-1.5B-Instruct}                                                                          \\ \midrule
SFT          & 0.094          & 0.193          & 0.129          & 0.210          & 0.024          & 0.130          \\
RAG          & 0.288          & 0.477          & 0.228          & 0.203          & 0.072          & 0.254          \\
PPO          & 0.327          & 0.527          & 0.256          & 0.242          & 0.184          & 0.307          \\
GRPO         & 0.293          & 0.480          & 0.202          & 0.213          & 0.120          & 0.261          \\
\hrpo (Ours) & \textbf{0.364} & \textbf{0.553} & \textbf{0.273} & \textbf{0.276} & \textbf{0.216} & \textbf{0.337} \\ \midrule
\multicolumn{7}{c}{Qwen2.5-3B-Instruct}                                                                            \\ \midrule
SFT          & 0.249          & 0.292          & 0.186          & 0.248          & 0.112          & 0.217          \\
RAG          & 0.348          & 0.544          & 0.255          & 0.226          & 0.080          & 0.291          \\
PPO          & 0.356          & 0.563          & 0.304          & 0.293          & 0.240          & 0.351          \\
GRPO         & \textbf{0.381} & 0.570          & 0.308          & 0.303          & 0.272          & 0.367          \\
\hrpo (Ours) & 0.378          & \textbf{0.593} & \textbf{0.316} & \textbf{0.318} & \textbf{0.296} & \textbf{0.380} \\ \bottomrule
\end{tabular}
\label{tab:qa-results}
\end{table}

\subsection{Evaluation on Knowledge Benchmarks}
We first evaluate \hrpo on five open‑domain and multi‑hop question answering (QA) datasets: Natural Questions (NQ), TriviaQA, HotpotQA, 2WikiMultiHopQA (2WikiMQA) and Bamboogle~\cite{ho2020constructing, joshi2017triviaqa, kwiatkowski2019natural, press2022measuring, yang2018hotpotqa}. For each query, we use the E5 embedding model~\cite{wang2022text} to retrieve the top‑3 Wikipedia documents as context (details presented in \Cref{sec:implementation}). Following~\cite{jin2025search}, we merge the NQ and HotpotQA training sets to train \hrpo models, and evaluate it on each dataset's evaluation split. The exact match results of \hrpo and baselines (including supervised fine-tuning (SFT), retrieval augmented generation (RAG)~\cite{lewis2020retrieval} and RL-based PPO~\cite{schulman2017proximal} and GRPO~\cite{shao2024deepseekmath}) for the 1.5B and 3B Qwen2.5 Instruct models~\cite{yang2024qwen2} are presented in \Cref{tab:qa-results}. We also include comparisons to several QA and RAG baselines using the larger Qwen2.5-7B-Instruct as backbone, including: direct inference (QA), chain-of-thought (CoT)~\cite{wei2022chain}, interleaving retrieval with CoT (IRCoT)~\cite{trivedi2023interleaving}, Search-o1~\cite{li2025search} and RAG~\cite{lewis2020retrieval}. For each block in \Cref{tab:qa-results}, we mark the best performance in bold for clarity.

Across all knowledge benchmarks, \hrpo delivers the strongest exact match (EM) scores with smaller Qwen models and rivals the much larger 7B baselines. In particular, we observe:
(1)~\hrpo reaches 0.380 EM with Qwen2.5-3B, outperforming the strongest 7B RAG baseline by 4.5\%. Similarly, \hrpo with the smaller 1.5B backbone scores an average of 0.337, achieving consistent gains and surpassing PPO by 3.0\%.
(2)~\hrpo consistently outperforms other RL-based methods. For example, \hrpo with both the 1.5B and 3B backbones surpasses the strongest RL baseline by 3.0\% and 1.3\% respectively; the only dataset both models perform similarly is NQ.
(3)~Interestingly, GRPO underperforms PPO by 4.6\% on the 1.5B backbone but outperforms it by 1.6\% on the 3B model, likely a consequence of sparser rewards and limited sampled trajectories with a smaller model.
(4)~RL-based methods perform on par with the best-performing RAG baseline, with \hrpo delivering the largest performance gains—particularly on terse, incomplete queries (NQ) and multi-hop questions (2WikiMQA)—while yielding modest improvements on one-hop datasets like TriviaQA.
Overall, these results demonstrate that combining retrieval augmentation with hybrid latent reasoning yields state-of-the-art knowledge performance under computation constraints, establishing \hrpo as a competitive alternative to both RL-based learning methods and larger retrieval augmented LLMs.

\begin{table}[t]
\centering
\caption{Evaluation performance of various larger LLMs and trained models on STEM benchmarks. The table presents accuracy scores on five datasets: GSM8k, MATH, MATH500, MMLU-ST and ARC-C. The upper block reports results for several few-shot baseline LLMs $\geq$ 7B, while the lower two blocks evaluate smaller Qwen models (1.5B and 3B) trained with different strategies.}
\begin{tabular}{@{}lcccccc@{}}
\toprule
                & GSM8k          & MATH           & MATH500        & MMLU-ST        & ARC-C          & Average        \\ \midrule
\multicolumn{7}{c}{Larger LLMs (Size $\geq$ 7B)}                                                                      \\ \midrule
DeepSeekMath-7B & 0.642          & 0.362          & 0.346          & 0.565          & 0.678          & 0.519          \\
Gemma-2-9B      & 0.707          & 0.377          & 0.364          & 0.651          & 0.682          & 0.556          \\
Qwen2.5-7B      & \textbf{0.854} & \textbf{0.498} & \textbf{0.464} & \textbf{0.723} & 0.637          & 0.635          \\
MAmmoTH2-7B     & 0.684          & 0.367          & 0.396          & 0.624          & 0.817          & 0.578          \\
MAmmoTH2-8B     & 0.704          & 0.358          & 0.732          & 0.642          & \textbf{0.822} & \textbf{0.652} \\ \midrule
\multicolumn{7}{c}{Qwen2.5-1.5B-Instruct}                                                                             \\ \midrule
SFT             & 0.560          & 0.300          & 0.302          & 0.403          & 0.602          & 0.433          \\
Distilled CoT   & 0.706          & 0.503          & -              & -              & -              & -              \\
PPO             & 0.694          & 0.507          & 0.518          & 0.566          & 0.715          & 0.600          \\
GRPO            & 0.711          & 0.502          & 0.524          & 0.562          & 0.737          & 0.607          \\
\hrpo (Ours)    & \textbf{0.720} & \textbf{0.518} & \textbf{0.536} & \textbf{0.569} & \textbf{0.742} & \textbf{0.617} \\ \midrule
\multicolumn{7}{c}{Qwen2.5-3B-Instruct}                                                                               \\ \midrule
SFT             & 0.670          & 0.348          & 0.360          & 0.454          & 0.474          & 0.461          \\
Distilled CoT   & 0.799          & 0.575          & -              & -              & -              & -              \\
PPO             & 0.819          & 0.597          & 0.604          & 0.582          & 0.811          & 0.682          \\
GRPO            & 0.834          & 0.602          & 0.604          & \textbf{0.601} & 0.814          & 0.691          \\
\hrpo (Ours)    & \textbf{0.845} & \textbf{0.613} & \textbf{0.630} & 0.590          & \textbf{0.820} & \textbf{0.700} \\ \bottomrule
\end{tabular}
\label{tab:math-results}
\end{table}

\subsection{Evaluation on STEM Benchmarks}
We also evaluate the performance of the proposed \hrpo on the reasoning-intensive STEM datasets: GSM8k, MATH, MATH500, MMLU-STEM (MMLU-ST) and ARC-Challenge (ARC-C)~\cite{cobbe2021training, hendrycks2021measuring, lightman2023let, hendrycks2020measuring, clark2018think}. \Cref{tab:math-results} reports the performance of \hrpo alongside fine-tuned baselines (SFT, SFT with distilled CoT from QwQ~\cite{qwq32b}) and RL baselines (PPO~\cite{schulman2017proximal} and GRPO~\cite{shao2024deepseekmath}) on the Qwen 2.5 1.5B and 3B Instruct models \cite{yang2024qwen2}. In addition, we select several larger LLMs ($\geq$ 7B in size) using few-shot CoT for comparison~\cite{yang2024qwen2, shao2024deepseekmath, yue2024mammoth2}. For GSM8k, we train on the training split, and for MATH and MATH500, we train on the MATH training split. For MMLU-ST and ARC-C, we train on the merged auxiliary MMLU and ARC-C training sets. Distilled CoT is only available for GSM8k and MATH due to dataset size constraints. We also highlight the best scores in each block in bold.

Across the five STEM benchmarks, \hrpo delivers the strongest results with compact Qwen backbones and could match the performance of much larger LLMs. Our key observations are:
(1)~SFT underperforms compared to distilled CoT and RL-based methods, suggesting the efficacy of RL with verifiable rewards on reasoning-intensive tasks.
(2)~With the 3B backbone, \hrpo achieves an average accuracy of 0.700, matching the best 7B baseline on four of the datasets. Even the 1.5B \hrpo averages at 0.617, outperforming the 7B leader on MATH by 2.0\%.
(3)~At 1.5B, \hrpo improves on the strongest alternative GRPO with notable boosts on MATH and MATH500 (1.6\% and 1.2\%), whereas the average gain narrows at 3B, implying that \hrpo is more beneficial for smaller models.
(4)~\hrpo registers the highest accuracies recorded for sub-7B models on MATH (0.613) and MATH500 (0.630), demonstrating the value of RL-based hybrid reasoning on challenging benchmarks.
Taken together, these results show that hybrid latent reasoning unlocks the power of much larger LLMs in compact backbones, proving the effectiveness of the proposed \hrpo.

\subsection{Analysis of \hrpo}
\label{sec:analysis}

\textbf{Different Strategies for Latent Reasoning.}
We compare different strategies to compute latent representations. Specifically, we use three methods to integrate hidden states into RL and train the 1.5B Qwen model on the MATH dataset. These variants are: (1)~hidden states, which use the final layer hidden states as the next input; (2)~interpolation, which employs interpolated embeddings
\begin{wrapfigure}[14]{r}{0.5\textwidth}
    \centering
    \includegraphics[width=\linewidth]{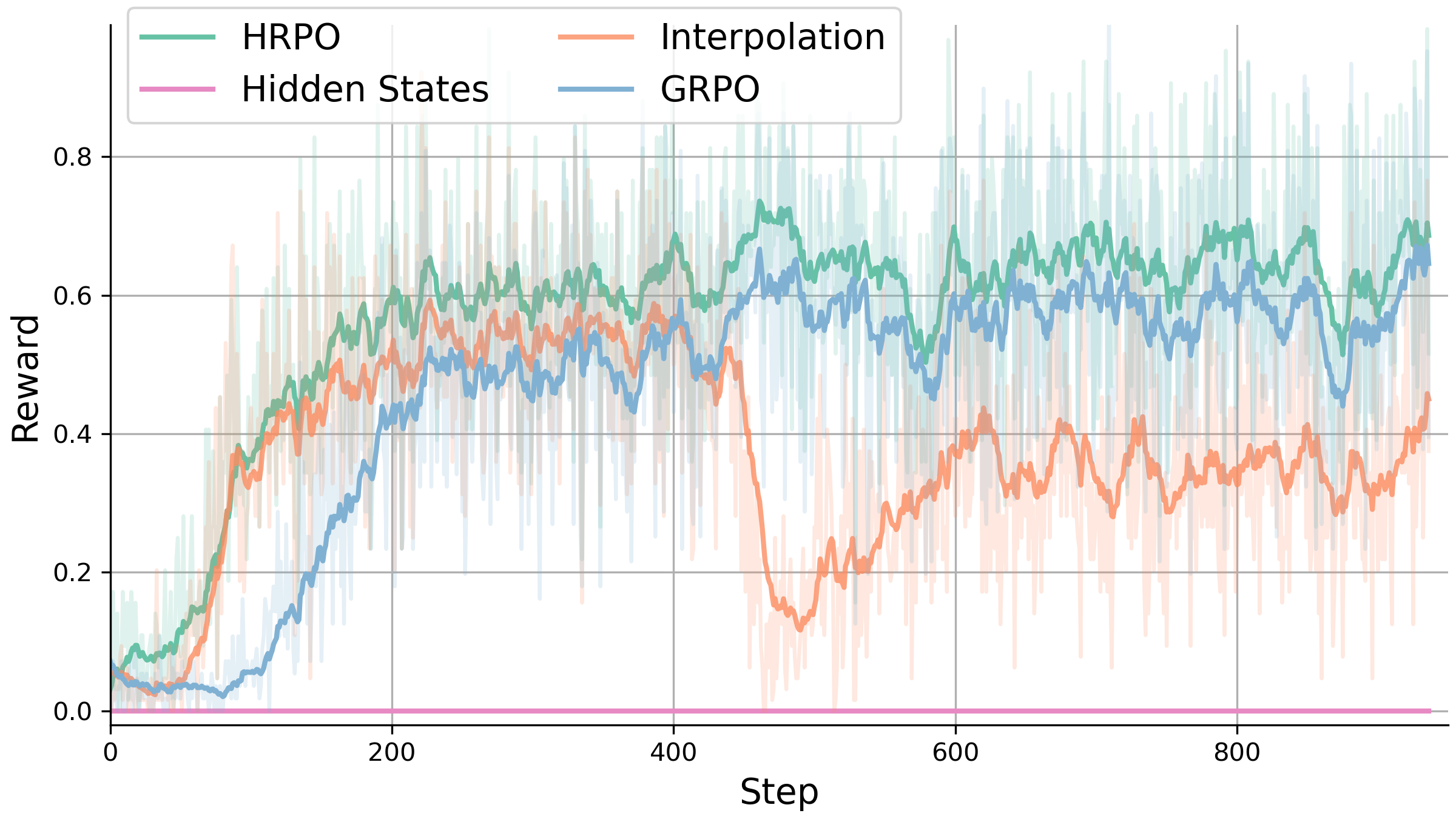}
    \caption{Reward on MATH for Qwen-2.5-1.5B using different latent reasoning strategies.}
    \label{fig:hidden-comparison}
\end{wrapfigure}
as defined in \Cref{eq:hidden_states}; and (3)~\hrpo, our hybrid latent reasoning in \Cref{eq:thinking_residual}. We visualize the exponential moving average (EMA) of rewards along with the GRPO baseline in \Cref{fig:hidden-comparison}. Due to the mismatch between hidden states and embeddings, using hidden states degrades generation and yields nonsensical rollouts with zero reward. Although interpolation performs similar to \hrpo for the first few hundred steps, the rewards eventually collapse and only slowly recover, likely because interpolation introduces excessive noise. We also provide a direct comparison between \hrpo and latent reasoning methods in \Cref{sec:additional-results}. Overall, our approach achieves superior training dynamics with faster convergence while maintaining stability comparable to GRPO, highlighting the efficacy of our hybrid design choice in \hrpo.

\begin{figure}[h]
    \centering
    \includegraphics[trim=0 0 0 0, clip, width=1.0\textwidth]{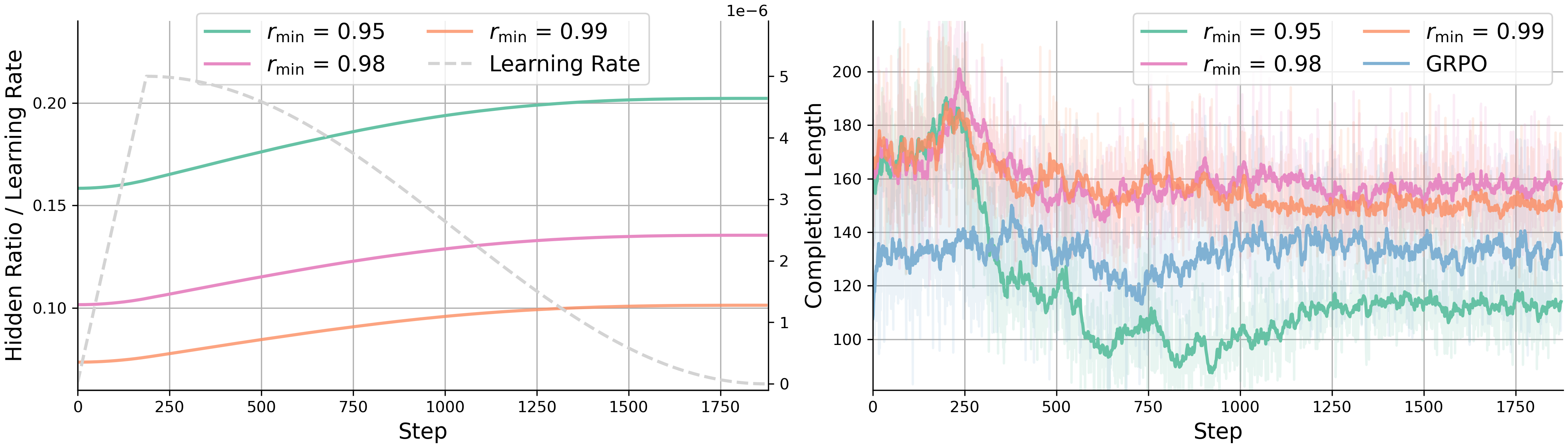}
    \caption{Hidden ratio with varying $r_{\mathrm{min}}$ in $\texttt{exp}(-c \cdot \texttt{softplus}(\Lambda))$ and learning rate. We visualize the hidden ratio and completion length for training runs with $r_{\mathrm{min}}$ from $[0.95, 0.98, 0.99]$.}
    \label{fig:hidden-ratio}
\end{figure}

\textbf{Ratio of Latent Representations.}
We track how the balance between discrete tokens and continuous latent representations shifts as LLMs learn to reason hybridly. Here, we train Qwen 1.5B on the knowledge task and visualize both the mean hidden ratios (i.e., $\sqrt{1-a_t^2}$) and completion lengths (along with GRPO) in \Cref{fig:hidden-ratio}. Across all runs, the hidden ratio increases steadily, even as the learning rate tapers off toward the end of training under a cosine schedule. In addition, completion lengths increase during the initial phase and later decline across all methods, with the drops most significant in \hrpo. Furthermore, setting $r_{\mathrm{min}} = 0.95$ leads to an interesting behavior where completion lengths substantially decrease—an effect not seen in the other variants\footnote{$r_\mathrm{min}$ is used to initialize $\Lambda$ such that $\texttt{exp}(-c \cdot \texttt{softplus}(\Lambda))$ is drawn uniformly from $[r_\mathrm{min}, 0.999]$.}. This may be because the hidden representations effectively capture historical context, thereby shortening completions while maintaining or even improving performance (see \Cref{tab:init-results}). As such, hybrid latent reasoning could be particularly effective when leveraging contextual information for reasoning.

\begin{table}[ht]
\small
\centering
\caption{Impact of $\Lambda$-initialization on \hrpo's performance across knowledge and STEM tasks.}
\begin{tabular}{@{}lcccccc@{}}
\toprule
\multicolumn{1}{c}{\multirow{2}{*}{Init Range}} & \multicolumn{6}{c}{Knowledge}                                                                       \\ \cmidrule(l){2-7} 
\multicolumn{1}{c}{}                            & NQ             & TriviaQA       & HotpotQA       & 2WikiMQA       & Bamboogle      & Average        \\ \midrule
{[}0.95 - 0.999{]}                              & \textbf{0.364} & \textbf{0.553} & \textbf{0.273} & 0.264          & 0.184          & 0.328          \\
{[}0.98 - 0.999{]}                              & 0.336          & 0.553          & 0.263          & \textbf{0.276} & \textbf{0.216} & \textbf{0.329} \\
{[}0.99 - 0.999{]}                              & 0.336          & 0.534          & 0.258          & 0.275          & 0.216          & 0.324          \\ \midrule
\multicolumn{1}{c}{\multirow{2}{*}{Init Range}} & \multicolumn{6}{c}{STEM}                                                                            \\ \cmidrule(l){2-7} 
                                                & GSM8k          & MATH           & MATH500        & MMLU-ST        & ARC-C          & Average        \\ \midrule
{[}0.95 - 0.999{]}                              & 0.705          & 0.516          & \textbf{0.536} & \textbf{0.569} & 0.735          & 0.612          \\
{[}0.98 - 0.999{]}                              & 0.703          & 0.509          & 0.532          & 0.563          & 0.732          & 0.608          \\
{[}0.99 - 0.999{]}                              & \textbf{0.720} & \textbf{0.518} & 0.526          & 0.567          & \textbf{0.742} & \textbf{0.614} \\ \bottomrule
\end{tabular}
\label{tab:init-results}
\end{table}

\textbf{Initialization of $\Lambda$ for Gating.}
Beyond hidden ratio, we examine how the initialization of $\Lambda$—which control the balance between latent features and token embeddings—affects \hrpo performance. Specifically, we initialize $\texttt{exp}(-c \cdot \texttt{softplus}(\Lambda))$ from $[r_{\mathrm{min}}, 0.999]$ and report the results on Qwen 1.5B in \Cref{tab:init-results}, where lowering $r_{\mathrm{min}}$ yields a higher initial hidden ratio. For the knowledge domain, performance improves as $r_{\mathrm{min}}$ decreases: the best average performance occurs at $r_{\mathrm{min}}=0.98$, and most individual datasets peak at $r_{\mathrm{min}}=0.95$. In contrast, the STEM benchmarks display a bimodal trend: performance rises when $r_{\mathrm{min}}$ is either lower or higher, but drops for the intermediate range $[0.98, 0.999]$. This pattern implies that the model profits from emphasizing either explicit token trajectories or latent representations, whereas a mid-level mix is sub-optimal. In summary, our results show that knowledge tasks benefit from lower $r_{\mathrm{min}}$, whereas optimal performance for STEM tasks arises from leaning toward either explicit token trajectories or latent representations.

\begin{figure}[t]
    \centering
    \includegraphics[trim=0 0 0 0, clip, width=1.0\textwidth]{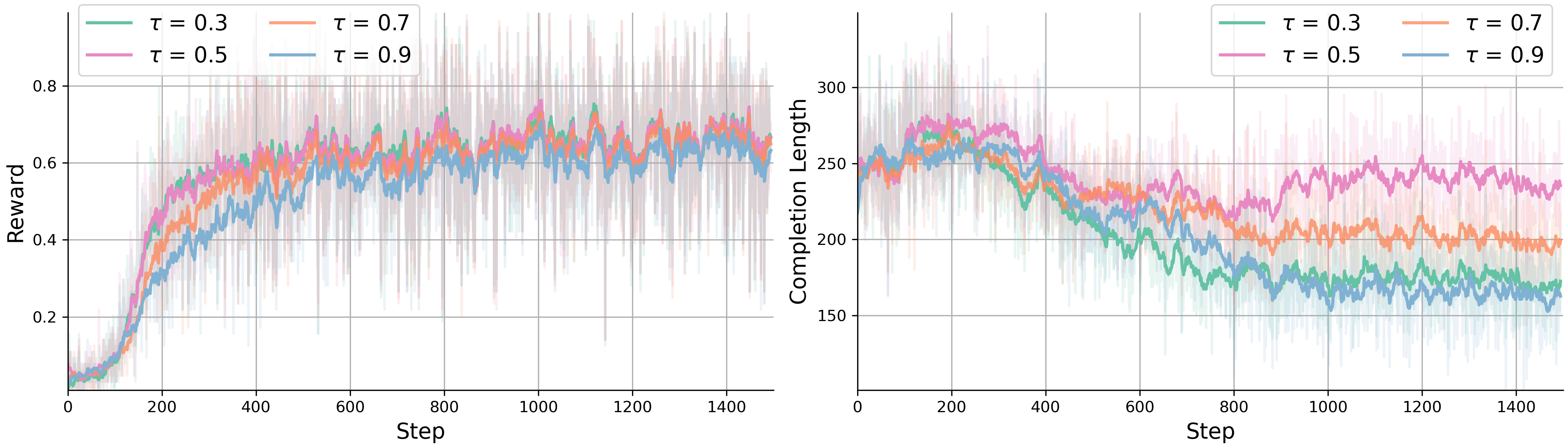}
    \caption{Sensitivity analysis for temperature $\tau$ in \Cref{eq:hidden_states}. We visualize the reward and completion length for training runs with different temperature selected from $[0.3, 0.5, 0.7, 0.9]$.}
    \label{fig:temperature}
\end{figure}

\textbf{Sensitivity of $\tau$ on Hybrid Reasoning.}
We further investigate the impact of temperature $\tau$ on \hrpo: lower $\tau$ values reduce noise but overemphasize top tokens, whereas larger $\tau$ spreads probability mass across more tokens. We explore $\tau \in \{ 0.3, 0.5, 0.7, 0.9 \}$ and present the rewards and completion lengths of the 1.5B Qwen model on MMLU in \Cref{fig:temperature}. The left panel indicates that $\tau = 0.3$ and $\tau = 0.5$ converge faster and reach the highest reward plateau, outperforming higher values ($\tau \geq 0.7$) and showing the benefits of a smaller $\tau$. Interestingly, the right panel reveals that both smaller and larger $\tau$ values shorten completion lengths, while $\tau = 0.5$ and $\tau = 0.7$ maintain longer generations. This may be because lower $\tau$ sharpens token distribution, yielding a confident latent vector that lets \hrpo finish quickly. In contrast, higher $\tau$ flattens the distribution and enhances informativeness, prompting the policy to extract answers in shorter rollouts. Overall, we find \hrpo to be robust across varuing $\tau$ selections, only completion length varies noticeably. Further analysis is in \Cref{sec:additional-results}.

\begin{figure}[h]
    \centering
    \includegraphics[trim=0.2cm 0.1cm 0.2cm 7.4cm, clip, width=1.0\textwidth]{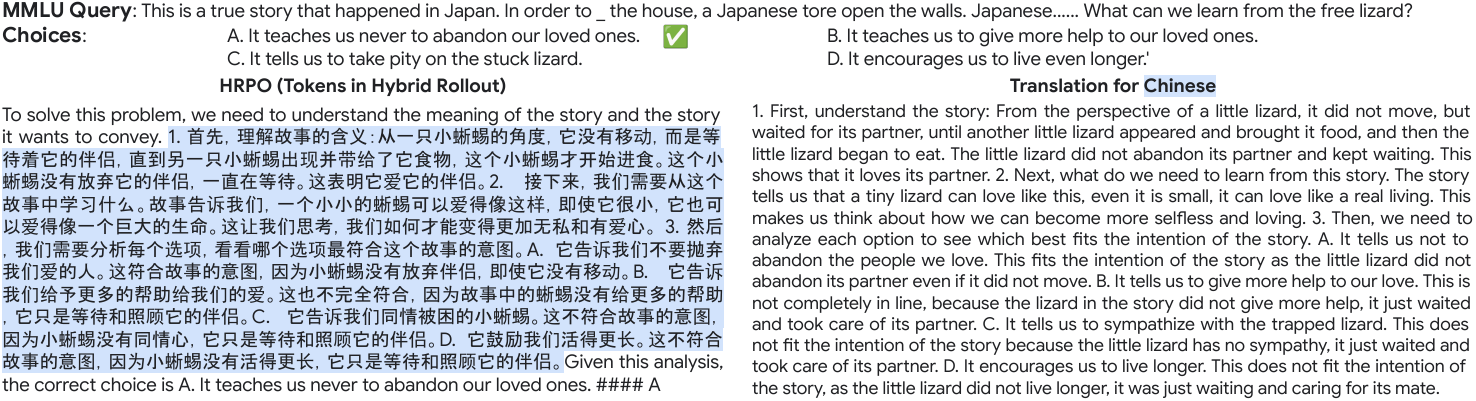}
    \caption{Example cross-lingual reasoning (English-Chinese) and its translation for \hrpo.}
    \label{fig:example-completion}
\end{figure}

\textbf{Hybrid Latent Reasoning Patterns.}
Finally, we highlight several intriguing reasoning patterns that emerge from \hrpo. First, the hybrid outputs show readable trajectories by interpreting the tokens even without any CoT supervision. Second, \hrpo exhibits cross-lingual patterns in some completions, fluidly integrating tokens from different languages, suggesting that latent representations can generalize across linguistic boundaries (see \Cref{fig:example-completion}). Moreover, the hybrid reasoning process often delivers compact yet accurate responses to simple or factual queries, where the model requires fewer decoding steps thanks to the richer context encoded in the hidden representations. These emergent patterns indicate that hybrid latent reasoning can improve both interpretability and efficiency over existing latent reasoning approaches. Further qualitative examples can be found in \Cref{sec:qualitative-analysis}.
\section{Conclusion}

In this work, we propose \hrpofull (\hrpo), a novel reinforcement learning (RL) framework that unifies discrete token sampling with continuous latent representations through a learnable gating mechanism. By gradually incorporating hidden features into sampled token embeddings, \hrpo incentivizes LLMs to refine their reasoning strategies hybridly. Extensive evaluations on knowledge and STEM benchmarks demonstrate that \hrpo outperforms both SFT and RL baselines, achieving consistent gains across diverse scenarios. Moreover, our analysis reveals that \hrpo not only ensures stable hybrid latent reasoning but also triggers intriguing reasoning patterns, showing its potential in reasoning-intensive settings and providing insights for RL-based continuous space learning. While promising, we recognize that \hrpo introduces additional computation overhead, the on-policy design may reduce large-scale training efficiency, and its continuous representations can be less transparent. Therefore, future work will aim to address these limitations by exploring simpler designs, off-policy extensions and advanced latent reasoning techniques to improve both the interpretability and efficiency of \hrpo.


\bibliographystyle{plainnat}   
\bibliography{reference}      


\appendix

\newpage

\section{Implementation}
\label{sec:implementation}
For hybrid latent reasoning, our plug-and-play component is by design compatible with any LLM architecture. We initialize its linear layers with a uniform distribution from $[-1/\sqrt{|H|}, 1/\sqrt{|H|}]$, where $|H|$ denotes the hidden state dimension. The gating parameter $\Lambda$ is selected such that the quantity $a^c = \texttt{exp}(-c \cdot \texttt{softplus}(\Lambda))$ is drawn uniformly from $[r_\mathrm{min}, 0.999]$, with the scalar constant fixed at $c=8$~\cite{de2024griffin}. Tuning $r_\mathrm{min}$ adjusts the initial fraction of hidden states involved in hybrid reasoning; a larger value increases the proportion of sampled token embeddings and can be helpful for enhancing generation quality during the initial training phase. Similarly, the temperature hyperparameter $\tau$ in \Cref{eq:hidden_states} can be tuned for optimal task performance, although \hrpo remains robust across a wide range of $\tau$ values. To efficiently train the LLMs with \hrpo, we patch the models with optimized kernel implementations\footnote{https://github.com/unslothai/unsloth} and employ low-rank adaptation (LoRA)~\cite{hu2021lora}. The default choice of hyperparameters are reported in \Cref{tab:hyperparameter} for \hrpo experiments. 

\begin{table}[h] 
    \centering
    \caption{Experiment hyperparameter settings.}
    \begin{tabular}{lccccc}
        \toprule
        Algorithm & \multicolumn{5}{c}{\texttt{HRPO}} \\ 
        Epochs & \multicolumn{5}{c}{\texttt{1}} \\
        Optimizer & \multicolumn{5}{c}{\texttt{AdamW 8bit}} \\
        Optimizer Momentum & \multicolumn{5}{c}{\texttt{$\beta_1$, $\beta_2$ = 0.9, 0.99}} \\
        Weight Decay & \multicolumn{5}{c}{\texttt{0.1}} \\
        Learning Rate & \multicolumn{5}{c}{\texttt{5e-6}} \\
        Learning Rate (Linear in \Cref{eq:thinking_residual}) & \multicolumn{5}{c}{\texttt{1e-4}} \\
        Learning Rate ($\Lambda$ in \Cref{eq:thinking_residual}) & \multicolumn{5}{c}{\texttt{1e-3}} \\
        \hrpo $\beta$ & \multicolumn{5}{c}{\texttt{0.005}} \\ 
        Max Gradient Norm & \multicolumn{5}{c}{\texttt{0.1}} \\
        Gradient Accumulation Step & \multicolumn{5}{c}{\texttt{4}} \\
        Group size $g$ in \hrpo & \multicolumn{5}{c}{\texttt{4 / 8}} \\
        Total Train Batch Size & \multicolumn{5}{c}{\texttt{32 / 64}} \\
        LR Scheduler & \multicolumn{5}{c}{\texttt{Cosine with Warmup}} \\
        Warmup Ratio & \multicolumn{5}{c}{\texttt{0.1}} \\
        Precision (WA) & \multicolumn{5}{c}{\texttt{BF16-mixed}} \\
        \midrule
        LoRA Modules & \multicolumn{5}{c}{\texttt{query, key, value, dense}} \\
        LoRA Rank & \multicolumn{5}{c}{\texttt{32}} \\
        LoRA $\alpha$ & \multicolumn{5}{c}{\texttt{64}} \\
        \bottomrule
    \end{tabular}
    \label{tab:hyperparameter}
\end{table}

The hyperparameters are selected empirically to balance efficiency and performance, and thanks to \hrpo's lightweight design and additional optimizations, our framework can run on a single GPU across diverse tasks. Additionally, we apply a larger weight-decay coefficient to (1)~enhance \hrpo training stability and (2)~encourage the gating towards incorporating more latent representations (since smaller positive $\Lambda$ values increase the hidden ratio $\sqrt{1-a_t^2}$). For simpler knowledge tasks and GSM8k, we fix the \hrpo group size at 4, which already delivers strong performance. For more challenging benchmarks, namely MATH, MATH500, MMLU-ST and ARC-C, we instead generate 8 hybrid completions for each query. As for prompt and completion lengths, we select them empirically based on our observations, and the selected values are summarized in \Cref{tab:length}.

\begin{table}[h] 
    \centering
    \caption{Experiment prompt / completion lengths.}
    \begin{tabular}{lccccc}
        \toprule
        Prompt / Completion Length for Knowledge Tasks & \multicolumn{5}{c}{\texttt{2048 / 512}} \\ 
        Prompt / Completion Length for GSM8k & \multicolumn{5}{c}{\texttt{512 / 512}} \\ 
        Prompt / Completion Length for MATH \& MATH500 & \multicolumn{5}{c}{\texttt{512 / 1024}} \\ 
        Prompt / Completion Length for MMLU-ST \& ARC-C & \multicolumn{5}{c}{\texttt{512 / 512}} \\ 
        \bottomrule
    \end{tabular}
    \label{tab:length}
\end{table}

For both training and evaluation, we build each prompt by prepending a system message that directs the LLM to perform step-by-step internal reasoning before generating its final answer. The user query is then appended, and the entire input is formatted with the model chat template. Different from prior work~\cite{guo2025deepseek, jin2025search}, we adopt the minimalist delimiter \texttt{\#\#\#\#} to separate the model's hybrid reasoning span from its final answer. This is because the delimiter tokenizes as a single unit, adding no length overhead while providing a clear signal to switch from hybrid latent reasoning to autoregressive answer generation. We also penalize repeated occurrences of the delimiter within the completion (by assigning 0 reward regardless answer correctness) to prevent the model from early termination of hybrid reasoning. We illustrate full prompts for different type of tasks, showing the system message and example queries in \Cref{fig:rag-prompt}, \Cref{fig:gsm8k-prompt} and \Cref{fig:mmlu-prompt}, respectively.

\begin{figure}[h]
\begin{prompt}{Example Prompt for Knowledge Tasks}
\texttt{<|im\_start|>system}

\texttt{A conversation between User and Assistant. The user asks a question, and the assistant solves it. The assistant first thinks about the reasoning process in the mind and then provides the user with the answer. The final answer is provided after the \#\#\#\# tag, i.e., \{reasoning process\} \#\#\#\# \{answer\}.<|im\_end|>}

\texttt{<|im\_start|>user}

\texttt{Context (which may or may not be relevant):}

\texttt{Clyde River (New South Wales)::::Clyde River (New South Wales) The...}

\texttt{Barwon River (New South Wales)::::River and Weir River (part of...}

\texttt{Taponga River::::Taponga River The Taponga River, an inland...}

\texttt{ }

\texttt{Question: What direction does the river that Austrolebias bellotti are found in flow?<|im\_end|>}

\texttt{<|im\_start|>assistant}
\end{prompt}
\caption{Example prompt for knowledge tasks, contexts are partially omitted due to space constraints.}
\label{fig:rag-prompt}
\end{figure}

\begin{figure}[h]
\begin{prompt}{Example Prompt for GSM8k / MATH / MATH500}
\texttt{<|im\_start|>system}

\texttt{A conversation between User and Assistant. The user asks a question, and the assistant solves it. The assistant first thinks about the reasoning process in the mind and then provides the user with the answer. The final answer is provided after the \#\#\#\# tag, i.e., \{reasoning process\} \#\#\#\# \{answer\}.<|im\_end|>}

\texttt{<|im\_start|>user}

\texttt{Natalia sold clips to 48 of her friends in April, and then she sold half as many clips in May. How many clips did Natalia sell altogether in April and May?<|im\_end|>}

\texttt{<|im\_start|>assistant}
\end{prompt}
\caption{Example prompt for GSM8k / MATH / MATH500 in \hrpo.}
\label{fig:gsm8k-prompt}
\end{figure}

\begin{figure}[h]
\begin{prompt}{Example Prompt for MMLU-ST / ARC-C}
\texttt{<|im\_start|>system}

\texttt{A conversation between User and Assistant. The user asks a question, and the assistant solves it. The assistant first thinks about the reasoning process in the mind and then provides the user with the answer. The final answer is provided after the \#\#\#\# tag, i.e., \{reasoning process\} \#\#\#\# \{answer\}.<|im\_end|>}

\texttt{<|im\_start|>user}

\texttt{Question: Two people are pushing a car. One person is pushing with a force of 450 N and the other person is pushing with a force of 300 N. What information is needed to determine the net force applied to the car by the people?}

\texttt{ }

\texttt{Options:}

\texttt{A. the direction of the road}

\texttt{B. the direction of the forces}

\texttt{C. the weight of the two people}

\texttt{D. the weight of the automobile<|im\_end|>}

\texttt{<|im\_start|>assistant}
\end{prompt}
\caption{Example prompt for MMLU-ST / ARC-C in \hrpo.}
\label{fig:mmlu-prompt}
\end{figure}

For each question in our knowledge-intensive QA setup, we embed the query with E5 embedding model~\cite{wang2022text}. The entire English Wikipedia 2020 dump is pre-encoded with the same model, after which we perform approximate nearest neighbor (ANN) search and select the three highest-scoring documents. These top-3 passages are concatenated to form the external context fed to the LLM, as illustrated in \Cref{fig:rag-prompt}. In our evaluation, we generate tokens using greedy decoding and compute latent representations according to \Cref{eq:hidden_states}, thereby ensuring the reproducibility of our results. For outcome-based reward and evaluation settings on knowledge tasks, we report exact match scores on val~/~test splits following~\cite{yue2024evidence, yue2025inference, jin2025search}. For mathematical (GSM8k, MATH and MATH500) and multiple-choice datasets (MMLU-ST and ARC-C), we follow~\cite{yue2024mammoth2} for post-processing and scoring. 

\section{Additional Results}
\label{sec:additional-results}

\textbf{Comparison to Latent Reasoning Methods.}
In addition to strong RL methods such as PPO and GRPO in our main experiments, we also benchmark the proposed \hrpo against additional latent reasoning baselines. Specifically, we evaluate \hrpo, Coconut and CODI on the GSM8K and MATH reasoning datasets, all using the 1.5B Qwen backbone. For Coconut, we train with its augmented CoT data (no MATH split is available), whereas for CODI we adopt the original datasets' CoT trajectories. The results are reported in \Cref{tab:latent-comparison}. We observe:
(1)~\hrpo achieves the best accuracy on both datasets, with 9.42\% and 23.63\% respective gains over the best performing latent reasoning baseline CODI.
(2)~Even compared to distilled CoT from a significantly larger model QwQ, \hrpo still scores consistent improvements on both datasets, showing the effectiveness of our hybrid latent reasoning.
(3)~Coconut lags behind on GSM8k, indicating limitations of latent reasoning by compressing CoT tokens, whereas CODI improves substantially with CoT SFT but still trails Distilled CoT and \hrpo.
Overall, \hrpo achieves the best performance against all baselines, demonstrating its consistent advantages over CoT distillation and prior latent reasoning methods.

\begin{table}[h]
\small
\centering
\caption{Performance comparison of \hrpo against alternative latent reasoning methods and distilled CoT baseline.}
\begin{tabular}{@{}lcccccccc@{}}
\toprule
                     & \multicolumn{2}{c}{\textbf{Coconut}} & \multicolumn{2}{c}{\textbf{CODI}}    & \multicolumn{2}{c}{\textbf{Distilled CoT}} & \multicolumn{2}{c}{\textbf{\hrpo}} \\ \cmidrule(l){2-3} \cmidrule(l){4-5} \cmidrule(l){6-7} \cmidrule(l){8-9} 
\multicolumn{1}{c}{} & GSM8k             & MATH             & GSM8k             & MATH             & GSM8k           & MATH                     & GSM8k           & MATH             \\ \midrule
Accuracy             & 0.315             & -                & 0.658             & 0.419            & 0.706           & 0.503                    & \textbf{0.720}  & \textbf{0.518}   \\ \bottomrule
\end{tabular}
\label{tab:latent-comparison}
\end{table}

\begin{table}[t]
\small
\centering
\caption{Impact of $\Lambda$-initialization on \hrpo's performance for the Qwen 3B backbone.}
\begin{tabular}{@{}lcccccc@{}}
\toprule
\multicolumn{1}{c}{\multirow{2}{*}{Init Range}} & \multicolumn{6}{c}{Knowledge}                                                                       \\ \cmidrule(l){2-7} 
\multicolumn{1}{c}{}                            & NQ             & TriviaQA       & HotpotQA       & 2WikiMQA       & Bamboogle      & Average        \\ \midrule
{[}0.95 - 0.999{]}                              & \textbf{0.845} & \textbf{0.613} & 0.622          & 0.576          & \textbf{0.820} & 0.695          \\
{[}0.98 - 0.999{]}                              & 0.842          & 0.600          & 0.614          & 0.585          & 0.813          & 0.691          \\
{[}0.99 - 0.999{]}                              & 0.838          & 0.606          & \textbf{0.630} & \textbf{0.590} & 0.817          & \textbf{0.696} \\ \midrule
\multicolumn{1}{c}{\multirow{2}{*}{Init Range}} & \multicolumn{6}{c}{STEM}                                                                            \\ \cmidrule(l){2-7} 
\multicolumn{1}{c}{}                            & GSM8k          & MATH           & MATH500        & MMLU-ST        & ARC-C          & Average        \\ \midrule
{[}0.95 - 0.999{]}                              & 0.367          & \textbf{0.593} & \textbf{0.316} & 0.311          & \textbf{0.296} & \textbf{0.377} \\
{[}0.98 - 0.999{]}                              & \textbf{0.378} & 0.588          & 0.311          & 0.298          & 0.296          & 0.374          \\
{[}0.99 - 0.999{]}                              & 0.375          & 0.584          & 0.309          & \textbf{0.318} & 0.288          & 0.375          \\ \bottomrule
\end{tabular}
\label{tab:init-results-3b}
\end{table}

\begin{figure}[h]
    \centering
    \includegraphics[trim=0 0 0 0, clip, width=1.0\textwidth]{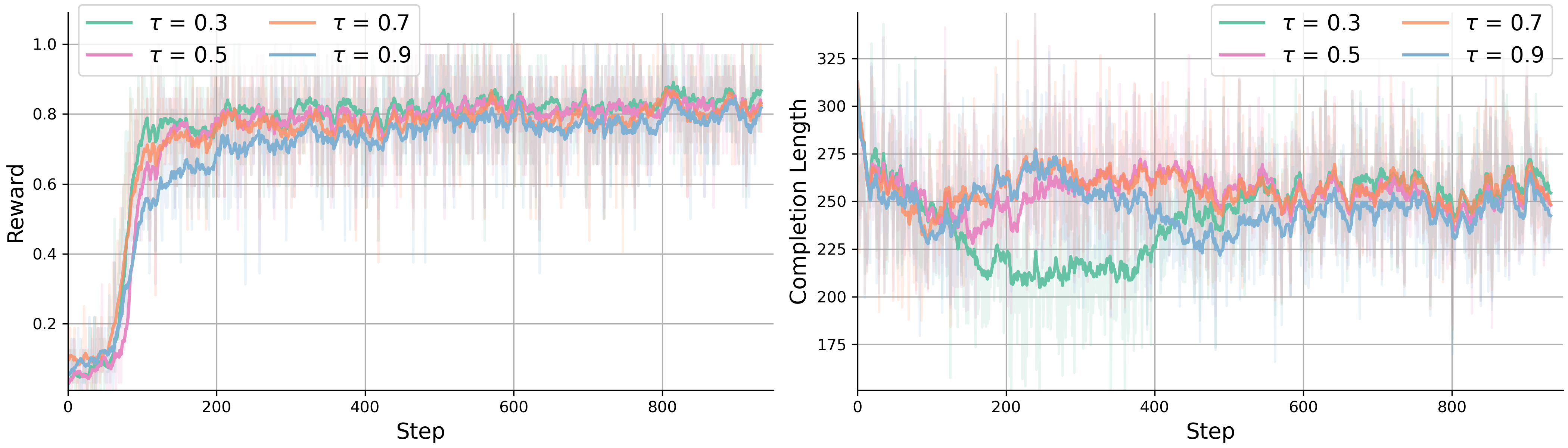}
    \caption{Reward and completion length for training runs with different temperature values on GSM8k using the Qwen 1.5B backbone.}
    \label{fig:temperature_gsm8k}
\end{figure}

\begin{figure}[h]
    \centering
    \includegraphics[trim=0 0 0 0, clip, width=1.0\textwidth]{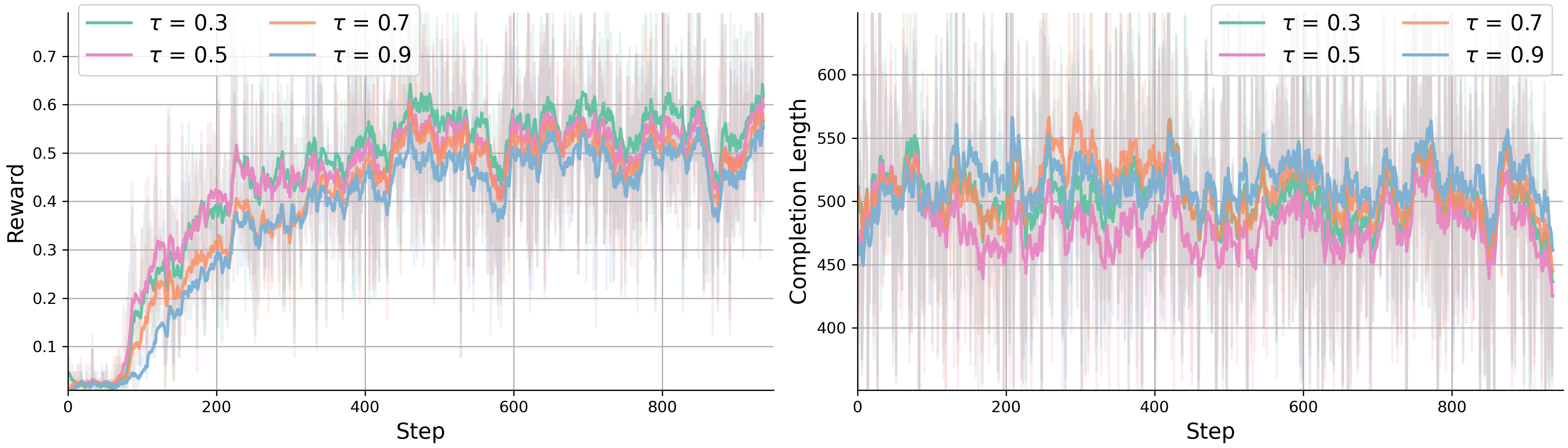}
    \caption{Reward and completion length for training runs with different temperature values on MATH using the Qwen 1.5B backbone.}
    \label{fig:temperature_math}
\end{figure}

\begin{figure}[h]
    \centering
    \includegraphics[trim=0 0 0 0, clip, width=1.0\textwidth]{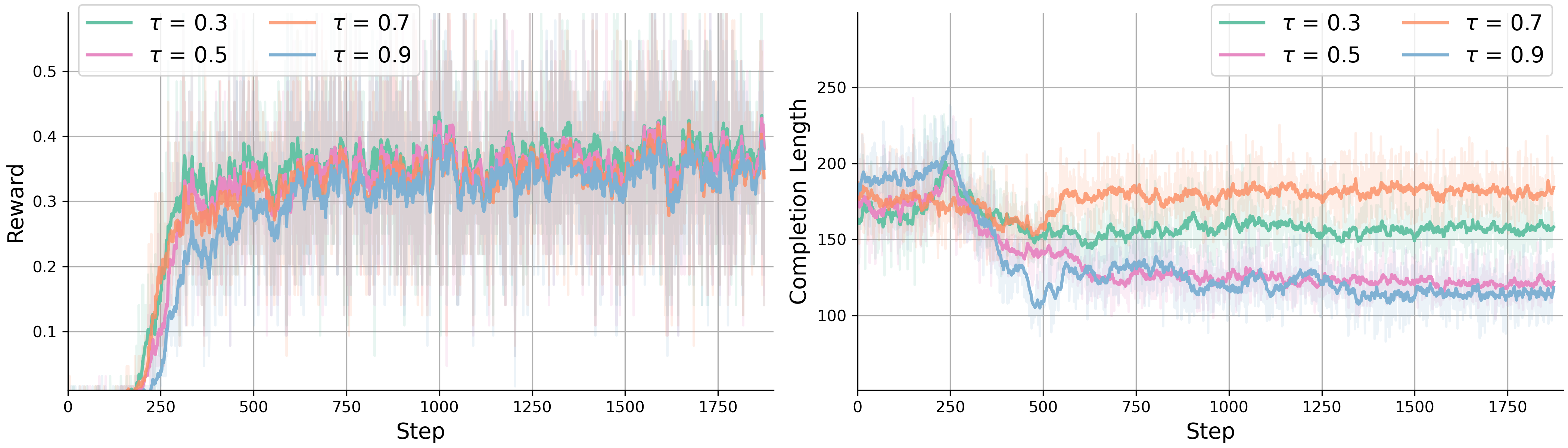}
    \caption{Reward and completion length for training runs with different temperature values on knowledge tasks using the Qwen 1.5B backbone.}
    \label{fig:temperature_rag}
\end{figure}

\textbf{Sensitivity Analysis for $\Lambda$ and $\tau$.}
In addition to the results reported in \Cref{tab:init-results}, we further present the performance of various $\Lambda$ initializations on the Qwen 3B model, as shown in \Cref{tab:init-results-3b}. Our observations echo the same trends on the 1.5B backbone: a smaller initial $r_{\mathrm{min}}$ consistently benefits both knowledge and STEM tasks. Moreover, performance peaks when $r_{\mathrm{min}}$ is selected either lower or higher, and drops slightly within the intermediate range of $[0.98, 0.999]$. We also examine the sensitivity of the $\tau$ hyperparameter used to construct the interpolated embeddings and present the corresponding results for both backbone models in \Cref{tab:tau-results}. The training rewards and completion lengths for GSM8k, MATH and the knowledge tasks are shown in \Cref{fig:temperature_gsm8k}, \Cref{fig:temperature_math} and \Cref{fig:temperature_rag}. We note that choosing $\tau$ in the range of 0.5 – 0.7 offers a reliable balance of efficiency and accuracy, as the performance often peaks around this interval for both backbone models. Overall, we find that \hrpo benefits from a smaller initial $r_{\mathrm{min}}$, which outperforms larger $r_{\mathrm{min}}$ settings and highlights the value of latent representations for complex reasoning. Moreover, \hrpo is robust to the choice of $\tau$, where the performance scores remain stable with only minor fluctuations at the extremes. 

\begin{table}[h]
\small
\centering
\caption{Impact of $\tau$ on \hrpo's performance for both backbone models.}
\begin{tabular}{@{}lcccccccc@{}}
\toprule
\multicolumn{1}{c}{Model}  & \multicolumn{4}{c}{Qwen2.5-1.5B}                                  & \multicolumn{4}{c}{Qwen2.5-3B}                                    \\ \cmidrule(l){2-5} \cmidrule(l){6-9} 
\multicolumn{1}{c}{$\tau$} & \textbf{0.3}   & \textbf{0.5}   & \textbf{0.7}   & \textbf{0.9}   & \textbf{0.3}   & \textbf{0.5}   & \textbf{0.7}   & \textbf{0.9}   \\ \midrule
GSM8k                      & 0.717          & \textbf{0.720} & 0.705          & 0.694          & 0.842          & 0.841          & \textbf{0.845} & 0.833          \\
MATH                       & \textbf{0.518} & 0.516          & 0.507          & 0.514          & 0.597          & 0.606          & \textbf{0.613} & 0.599          \\
MATH500                    & 0.522          & \textbf{0.536} & 0.532          & 0.524          & 0.622          & 0.614          & 0.622          & \textbf{0.630} \\
MMLUST                     & 0.561          & \textbf{0.569} & 0.559          & 0.567          & 0.577          & \textbf{0.590} & 0.574          & 0.580          \\
ARC-C                      & 0.735          & 0.741          & \textbf{0.742} & 0.724          & \textbf{0.820} & 0.817          & 0.809          & 0.808          \\ \midrule
NQ                         & 0.320          & 0.336          & 0.317          & \textbf{0.364} & \textbf{0.378} & 0.375          & 0.373          & 0.363          \\
TQ                         & 0.524          & 0.534          & \textbf{0.553} & 0.553          & 0.588          & \textbf{0.593} & 0.578          & 0.578          \\
HotpotQA                   & 0.263          & 0.260          & 0.252          & \textbf{0.273} & 0.311          & \textbf{0.316} & 0.309          & 0.306          \\
2Wiki                      & \textbf{0.276} & 0.272          & 0.264          & 0.244          & \textbf{0.318} & 0.311          & 0.297          & 0.293          \\
Bamboogle                  & 0.216          & 0.216          & \textbf{0.216} & 0.176          & 0.296          & 0.288          & \textbf{0.296} & 0.280          \\ \bottomrule
\end{tabular}
\label{tab:tau-results}
\end{table}

\begin{figure}[h]
    \centering
    \includegraphics[trim=0 0 0 0, clip, width=1.0\textwidth]{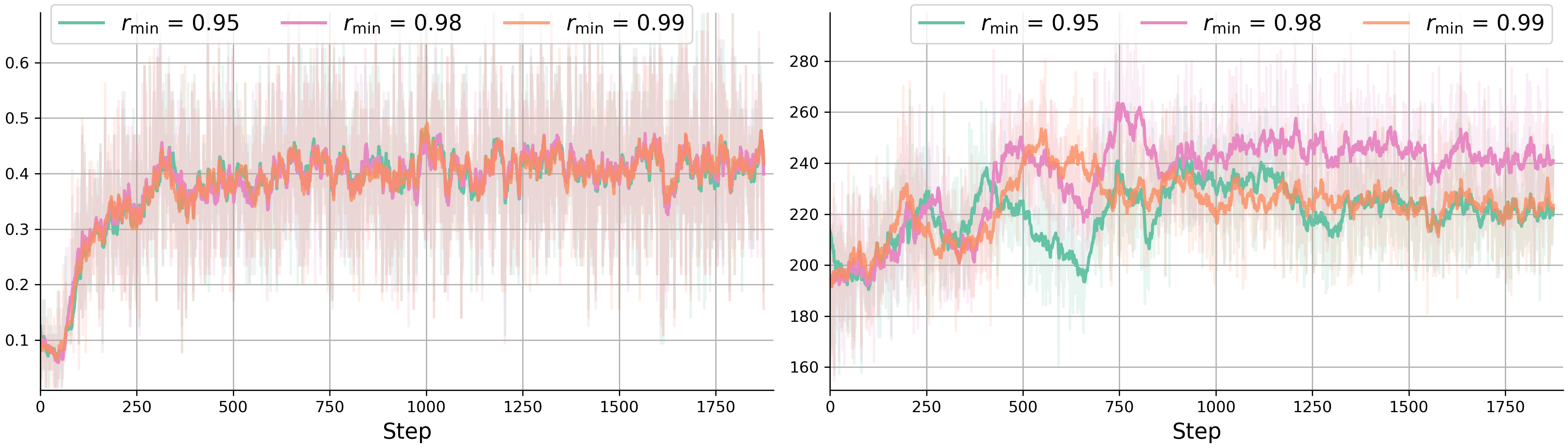}
    \caption{Reward and completion length for training runs with varying initial $r_\mathrm{min}$ on knowledge tasks using the Qwen 3B backbone.}
    \label{fig:reward_length_rag3b}
\end{figure}

\begin{figure}[h]
    \centering
    \includegraphics[trim=0 0 0 0, clip, width=1.0\textwidth]{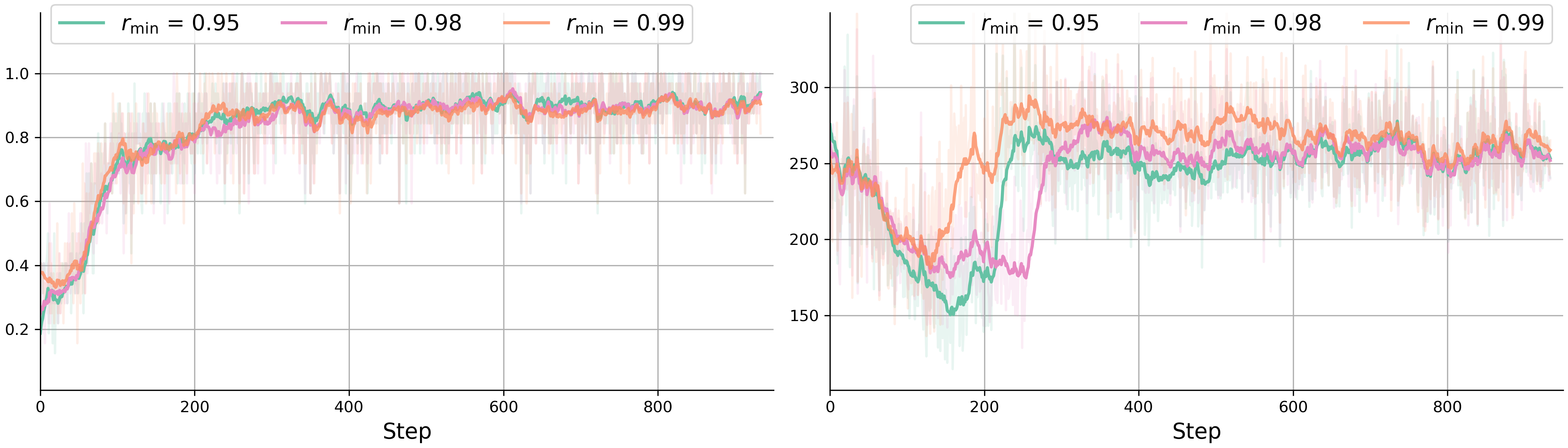}
    \caption{Reward and completion length for training runs with varying initial $r_\mathrm{min}$ on GSM8k using the Qwen 3B backbone.}
    \label{fig:reward_length_gsm8k3b}
\end{figure}

\begin{figure}[h]
    \centering
    \includegraphics[trim=0 0 0 0, clip, width=1.0\textwidth]{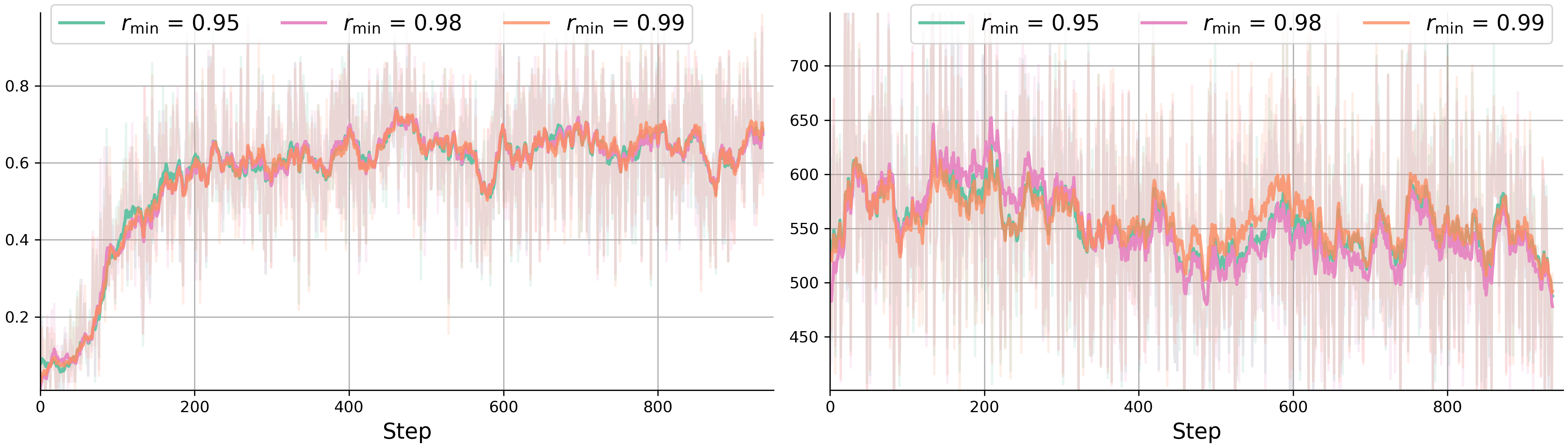}
    \caption{Reward and completion length for training runs with varying initial $r_\mathrm{min}$ on MATH using the Qwen 3B backbone.}
    \label{fig:reward_length_math3b}
\end{figure}

\begin{figure}[h]
    \centering
    \includegraphics[trim=0 0 0 0, clip, width=1.0\textwidth]{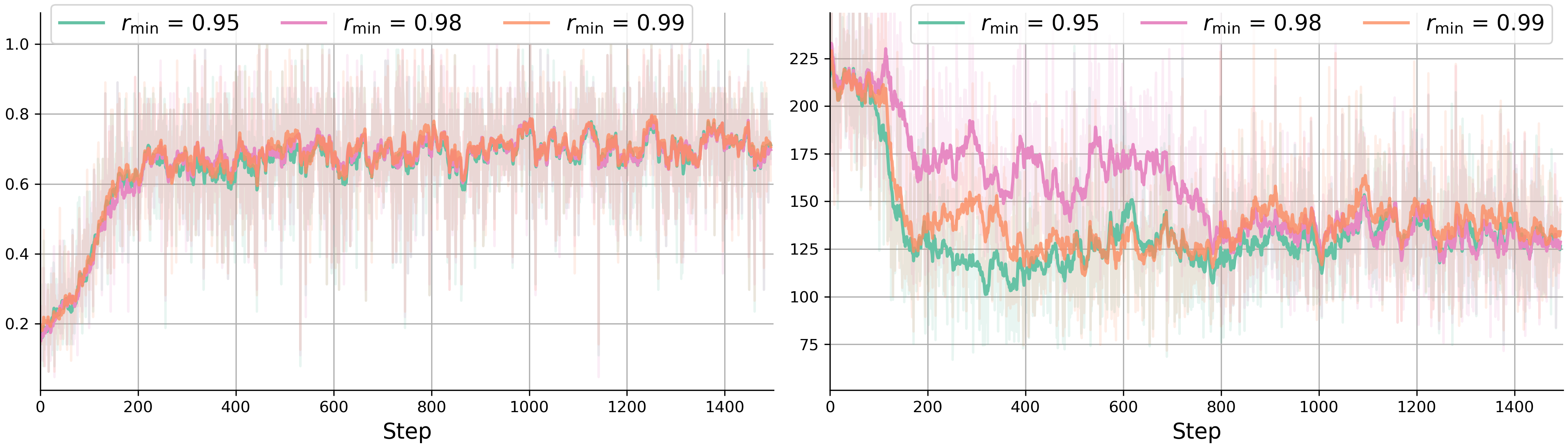}
    \caption{Reward and completion length for training runs with varying initial $r_\mathrm{min}$ on MMLU-ST / ARC-C using the Qwen 3B backbone.}
    \label{fig:reward_length_mmlu3b}
\end{figure}

\textbf{Additional Analysis for $\Lambda$ Initialization.}
We further provide an expanded analysis of how varying $r_{\mathrm{min}}$ in the initialization of $\Lambda$ affects training dynamics with the larger Qwen 3B backbone. Figures \Cref{fig:reward_length_rag3b}, \Cref{fig:reward_length_gsm8k3b}, \Cref{fig:reward_length_math3b} and \Cref{fig:reward_length_mmlu3b} plot the reward and completion length curves for the knowledge tasks, GSM8k, MATH and MMLU-ST / ARC-C respectively. Overall, our findings here echo the observations in \Cref{sec:analysis}: different $r_{\mathrm{min}}$ values exhibit similarly high training stability and preserve the LLM's generative capabilities, but selecting a smaller $r_{\mathrm{min}}$ (i.e., a larger initial hidden ratio) generally accelerates convergence and shortens generated completions. Nevertheless, these benefits are less pronounced for the 3B backbone than for the 1.5B counterpart, which we attribute to the fewer update steps and trainable parameters in \hrpo. In summary, our analysis shows that \hrpo preserves stable training dynamics and effectively leverages LLMs' intrinsic reasoning patterns across $r_{\mathrm{min}}$ values; moreover, choosing a smaller $r_{\mathrm{min}}$ further enhances convergence and yields shorter generated sequences, which can be especially beneficial for smaller-scale LLMs.

\textbf{Statistical Significance Analysis on the Improvements of \hrpo.}
In our main experiments, we follow the standard practice of using greedy decoding for pass@1 evaluation, ensuring our results are easy to evaluate and reproducible. To evaluate the significance of the performance gains of \hrpo, we conduct additional sampling-based evaluations on the STEM tasks, which exhibit greater variance compared to greedy decoding. Averaged results are presented in \Cref{tab:significance-test}, with statistically significant outcomes (paired t-test, $p < 0.05$) highlighted in bold. These results show that \hrpo consistently outperforms PPO and GRPO across both backbones on all benchmark datasets. For the 1.5B backbone, t-tests confirm these gains are statistically significant in three out of five tasks. The improvements are even more pronounced with the 3B model, which achieves an average gain of +1.4\% and shows statistical significance in four out of five comparisons. These findings demonstrate that our hybrid-RL framework, \hrpo, not only delivers reliable performance gains over established baselines but also does so with high statistical confidence across the majority of STEM tasks.

\begin{table}[ht]
\small
\centering
\caption{Significance test on \hrpo's performance improvements.}
\begin{tabular}{@{}lccccc@{}}
\toprule
                     & \multicolumn{5}{c}{Qwen2.5-1.5B}                                                   \\ \cmidrule(l){2-6} 
                     & GSM8k          & MATH           & MATH500        & MMLU-ST        & ARC-C          \\ \midrule
PPO                  & 0.701          & 0.505          & 0.511          & 0.551          & 0.716          \\
GRPO                 & 0.710          & 0.510          & 0.512          & 0.554          & 0.722          \\
\hrpo                & 0.712          & \textbf{0.515} & 0.517          & \textbf{0.565} & \textbf{0.731} \\ \midrule
                     & \multicolumn{5}{c}{Qwen2.5-3B}                                                     \\ \cmidrule(l){2-6} 
                     & GSM8k          & MATH           & MATH500        & MMLU-ST        & ARC-C          \\ \midrule
PPO                  & 0.825          & 0.597          & 0.600          & 0.574          & 0.802          \\
GRPO                 & 0.827          & 0.595          & 0.599          & 0.577          & 0.808          \\
\hrpo                & \textbf{0.838} & \textbf{0.606} & \textbf{0.609} & \textbf{0.585} & \textbf{0.815} \\ \bottomrule
\end{tabular}
\label{tab:significance-test}
\end{table}

\section{Qualitative Analysis}
\label{sec:qualitative-analysis}

\begin{figure}[h]
    \centering
    \includegraphics[trim=0 0 11.5cm 10cm, clip, width=1.0\textwidth]{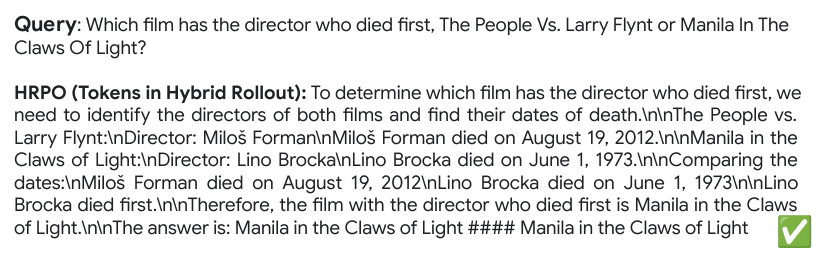}
    \caption{Correct reasoning example 1 in \hrpo.}
    \label{fig:correct0}
\end{figure}

\begin{figure}[h]
    \centering
    \includegraphics[trim=0 0 11.5cm 8cm, clip, width=1.0\textwidth]{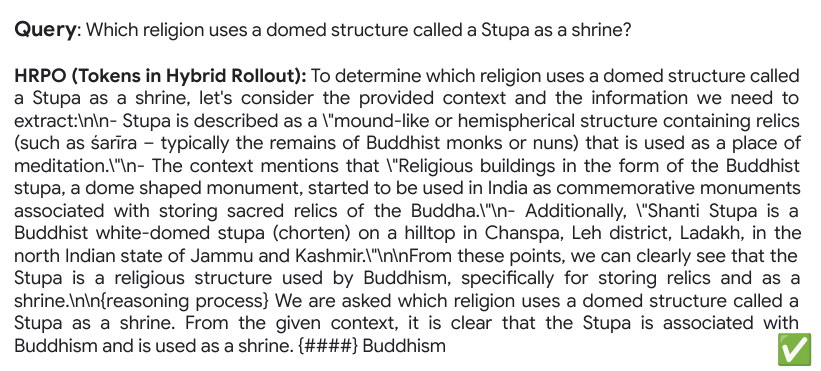}
    \caption{Correct reasoning example 2 in \hrpo.}
    \label{fig:correct1}
\end{figure}

\begin{figure}[h]
    \centering
    \includegraphics[trim=0 0 11.5cm 10cm, clip, width=1.0\textwidth]{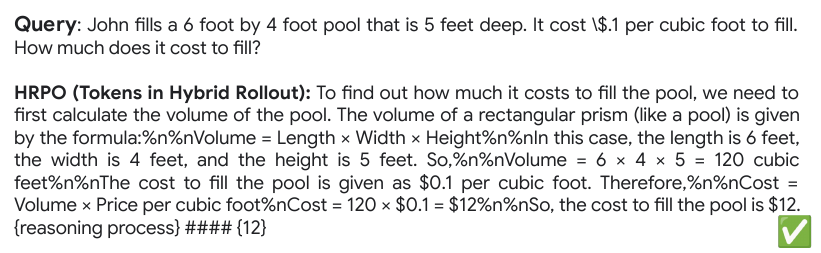}
    \caption{Correct reasoning example 3 in \hrpo.}
    \label{fig:correct2}
\end{figure}

\begin{figure}[h]
    \centering
    \includegraphics[trim=0 0 11.5cm 10cm, clip, width=1.0\textwidth]{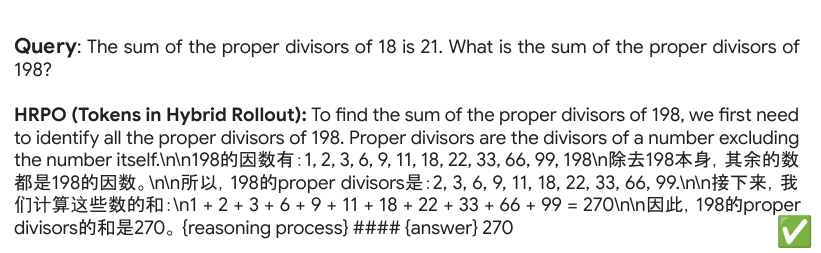}
    \caption{Correct reasoning example 4 in \hrpo.}
    \label{fig:correct3}
\end{figure}

\begin{figure}[h]
    \centering
    \includegraphics[trim=0 0 11.5cm 8cm, clip, width=1.0\textwidth]{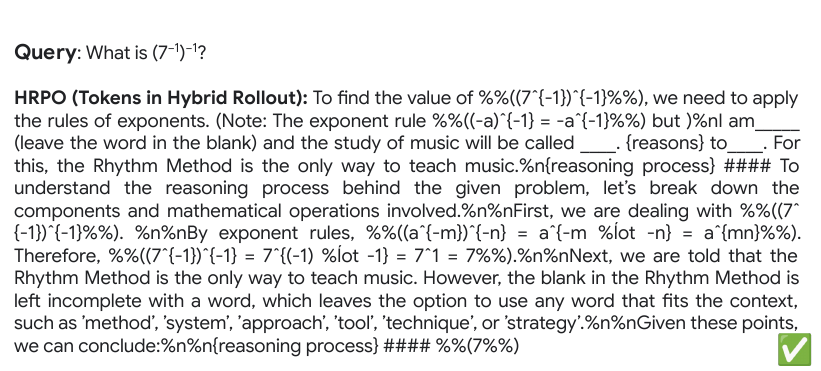}
    \caption{Correct reasoning example 5 in \hrpo.}
    \label{fig:correct4}
\end{figure}

\begin{figure}[h]
    \centering
    \includegraphics[trim=0 0 11.5cm 11cm, clip, width=1.0\textwidth]{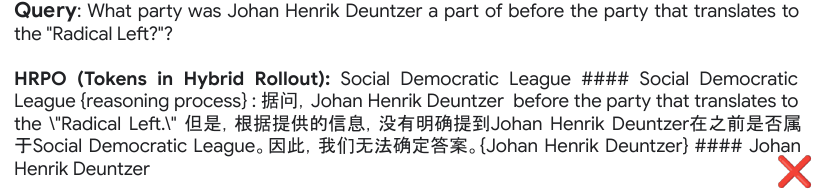}
    \caption{Mistaken reasoning example 1 in \hrpo.}
    \label{fig:wrong0}
\end{figure}

\begin{figure}[h]
    \centering
    \includegraphics[trim=0 0 11.5cm 11cm, clip, width=1.0\textwidth]{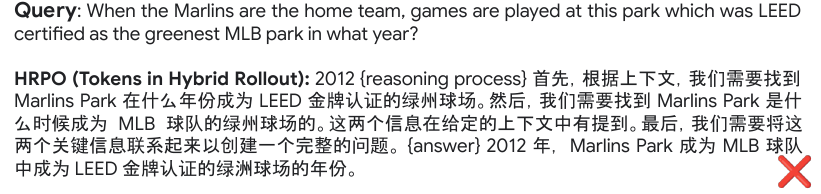}
    \caption{Mistaken reasoning example 2 in \hrpo.}
    \label{fig:wrong1}
\end{figure}

\begin{figure}[h]
    \centering
    \includegraphics[trim=0 0 11.5cm 8cm, clip, width=1.0\textwidth]{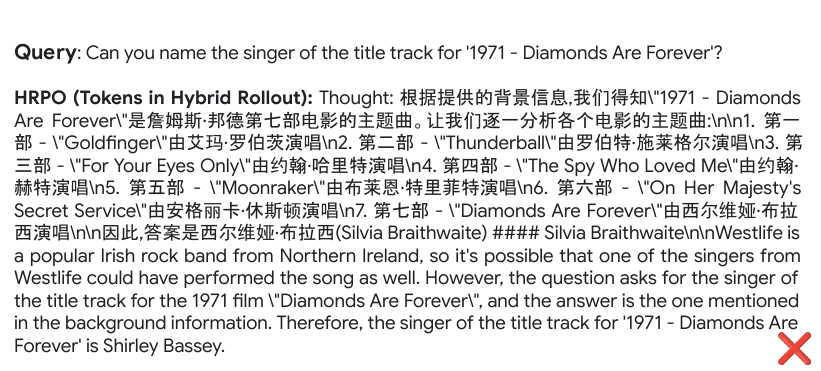}
    \caption{Mistaken reasoning example 3 in \hrpo.}
    \label{fig:wrong2}
\end{figure}

\begin{figure}[h]
    \centering
    \includegraphics[trim=0 0 11.5cm 8cm, clip, width=1.0\textwidth]{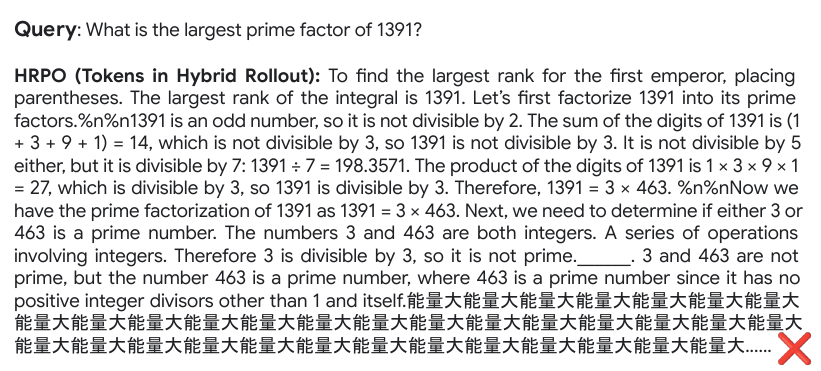}
    \caption{Mistaken reasoning example 4 in \hrpo.}
    \label{fig:wrong3}
\end{figure}

\begin{figure}[h]
    \centering
    \includegraphics[trim=0 0 11.5cm 0, clip, width=1.0\textwidth]{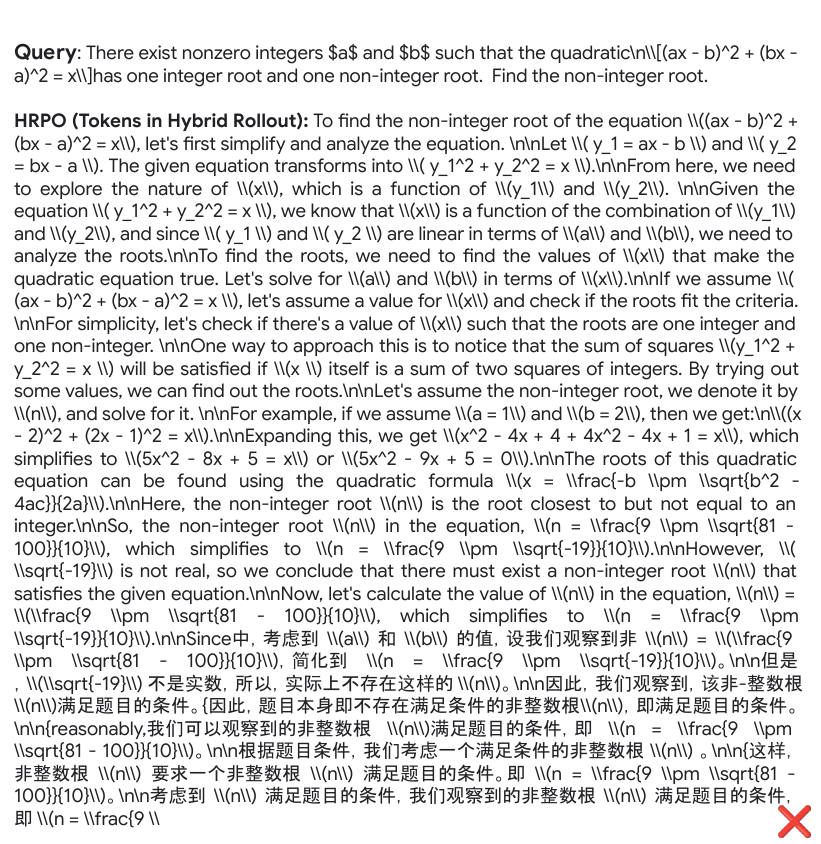}
    \caption{Mistaken reasoning example 5 in \hrpo.}
    \label{fig:wrong4}
\end{figure}

To further highlight \hrpo's reasoning patterns, we present additional qualitative examples. Each example provides the reasoning trace by decoding the sampled tokens from the hybrid reasoning process, and we include both successful and erroneous cases across different tasks in the following. The correct examples are provided in \Cref{fig:correct0}, \Cref{fig:correct1}, \Cref{fig:correct2}, \Cref{fig:correct3}, \Cref{fig:correct4}, where as the mistakes are provided in \Cref{fig:wrong0}, \Cref{fig:wrong1}, \Cref{fig:wrong2}, \Cref{fig:wrong3}, \Cref{fig:wrong4}, we show the raw strings and omit the options~/~contexts in the examples due to space constraints.

From these examples, we identify four reasoning patterns that can lead to correct answers: 
(1)~Purely English reasoning with coherent trajectories (Figs. \Cref{fig:correct0} and \Cref{fig:correct1}), a pattern commonly observed in LLM reasoning outputs.
(2)~Predominantly English reasoning punctuated by rare tokens (e.g., \%n rather than $\backslash$n), as shown in \Cref{fig:correct2}).
(3)~Cross-lingual reasoning that interweaves multiple languages (English and Chinese in \Cref{fig:correct3}). 
(4)~Reasoning with many uncommon tokens and atypical steps, yet still arriving at the correct answer (\Cref{fig:correct4}).
These latter three patterns are rarely observed in standard reasoning LLMs but are more prevalent in \hrpo trained models, demonstrating that \hrpo can enhance reasoning by leveraging LLMs' intrinsic generative capabilities across different languages and token types, thereby delivering improvements across diverse scenarios.

As for reasoning errors, we also identify several common patterns:
(1)~Cross-lingual mistakes arising from limited parametric or contextual knowledge, as in \Cref{fig:wrong0} and \Cref{fig:wrong1}.
(2)~Correct answers that violate the predefined format and thus receive a zero score (\Cref{fig:wrong2}).
(3)~Repetitive loops that continue until the response hits the maximum completion length (\Cref{fig:wrong3}).
(4)~Cross-lingual reasoning that is nonetheless truncated by the length limit (\Cref{fig:wrong4}).
Overall, these patterns indicate that, while \hrpo effectively integrates discrete and latent representations in its internal reasoning process, it may be further enhanced through refined output formatting (e.g., potentially with a format reward), extended optimization schedules with conservative learning, increased model parameters, and longer context~/~generation capabilities, pointing to promising directions for future research.

\end{document}